\documentclass[letterpaper]{article}
\usepackage[preprint]{aaai2027}
\usepackage[hyphens]{url}
\usepackage{graphicx}
\usepackage{xcolor}
\usepackage{natbib}
\usepackage{amsmath,amssymb}
\usepackage{caption}
\usepackage{subcaption}
\usepackage{multirow}
\usepackage{algorithm}
\usepackage{algpseudocode}
\newcommand{\best}[1]{\textcolor{red}{#1}}
\newcommand{\second}[1]{\textcolor{blue}{#1}}


\usepackage{xspace}
\newcommand{\m}{\texttt{ZeroHAT}\xspace}

\title{\m: Behavior-Conditioned Zero-Shot Human Activity Trace Generation}
\author{
    Rongchao Xu,
    Dahai Yu,
    Lin Jiang,
    Guang Wang
}

\affiliations{
    Florida State University\\
    Tallahassee, FL, USA\\
    \{rx21a, dahai.yu, lj23d, guang.wang\}@fsu.edu
}

\begin{document}

\maketitle

\begin{abstract}
Human activity traces (HATs) record individuals' timestamped visits to points of interest (POIs) and are essential for applications such as mobility prediction, urban simulation, and location-based services. However, accessing large-scale HATs is challenging due to high collection costs and privacy concerns. Synthetic HAT generation offers a promising way to make such data available and has attracted growing interest from both industry and academia. 
Although many efforts have been devoted to this topic, most of them rely on real data from a region to generate synthetic data for the same region, which is infeasible for the many regions where real HATs are unavailable.
To fill this gap, we propose \m, a behavior-conditioned framework that generates synthetic HATs for a target region in a zero-shot manner by transferring behavioral patterns learned from real HATs in source regions and adapting them with publicly available contextual information about the target region. \m has three key novel components: (i) a multidimensional consistency-aware intent extractor summarizes the generated prefix and extracts temporal, semantic, and relative spatial consistency intents;
(ii) a cross-region behavioral cloning module learns approximately region-invariant actions for revisiting recent, frequent, or anchor POIs; and (iii) a behavior-conditioned activity realization module constructs a dynamic action--POI graph to ground the action distribution onto compatible target POIs, which requires no explicit cross-region POI alignment, and a consistency-guided product-of-experts integrates grounded behavior, prefix consistency, and identity-free transition evidence to generate the next timestamped activity. We evaluate \m on a ten-city benchmark, where extensive experiments show that \m achieves $4.5$--$6.4\times$ the normalized downstream utility of the strongest baseline and improves average fidelity by $15.6\%$--$40.8\%$ across target regions.
% The source code is available at \url{https://anonymous.4open.science/r/ZeroHAT-207F}.
\end{abstract}

% Existing generators learn rich spatiotemporal and semantic dependencies but require target-region HATs and rely on region-specific POI representations and transition patterns.
% Because individual HATs are available in only a subset of regions, we ask whether source-region HATs can support generation in unseen regions without target traces.
% Recent cross-region and zero-shot mobility generators reduce target-data dependence, but primarily model coordinate-, grid-, or zone-level movement rather than timestamped semantic activities over target-private POI vocabularies.
\section{Introduction}

Human activity traces (HATs) are timestamped sequences of visits to semantically labeled points of interest (POIs), capturing when, where, and for what purposes people conduct activities in urban environments.
They support a wide range of real-world applications, including mobility prediction, next-POI recommendation, transportation planning, urban simulation, and location-based services.

However, accessing large-scale individual HATs is challenging because collecting such data is costly and time-consuming, while the sensitive location and behavioral information they contain raises substantial privacy concerns that often restrict data access and sharing.
These limitations have motivated growing interest in \emph{synthetic HAT generation}, which seeks to produce realistic HATs while reducing dependence on directly collected individual records.
Although various methods have been proposed for HAT generation, most of them focus on \emph{single-region generation}~\citep{geogen2026,gong2025stcdm,xu2026synhat}: they are trained and evaluated on HATs from the same region. As a result, these models may not be directly applied to a new region with a distinct set of POIs and human behaviors, especially when target-region HATs are unavailable for retraining or adaptation.
Some recent efforts generate trajectories for unseen regions, but their outputs are coarse-grained grid sequences rather than semantically rich HATs~\citep{mobta2026}.
Other works~\citep{cola2024} support cross-region transfer but still rely on target-specific parameters and frequency priors derived from target trajectories.

To fill this gap, we study \emph{zero-shot HAT generation}, which generates synthetic fine-grained HATs for a target region with authentic HATs from other source regions instead of the target region. This task poses two fundamental challenges.
First, capturing inherent behavioral patterns from HATs is essential for maintaining spatiotemporal and semantic coherence in synthetic HATs, yet it is non-trivial due to irregular activity intervals and diverse behavioral choices, which cause uncertainty about when and where the next activity will occur.
Second, transferring complex individual behaviors from source regions to an unseen target region is also challenging due to differences in POI distributions, spatial layouts, category supply, popularity, and accessibility across regions.

 To address these challenges, we propose \m, a behavior-conditioned framework that generates synthetic HATs for a target region in a zero-shot manner by learning transferable behavioral patterns from authentic HATs in source regions and conditioning them with already available contextual information (e.g., POI catalogs) about the target region.
First, to capture the spatiotemporal and semantic characteristics of the generated HAT prefix, we design a multidimensional consistency-aware intent extractor.
It takes region-agnostic attributes of each activity as input and employs a Mamba backbone to extract temporal, semantic, and relative spatial consistency intents from the evolving HAT state.
Second, driven by the finding that many behavioral structures are transferable across regions with different contexts, such as returning to recently visited, frequently visited, or regularly visited locations (e.g., repeatedly returning home), we introduce cross-region behavioral cloning over a region-invariant action space, where a policy conditioned on the consistency intents and structured prefix memory learns a behavior-action distribution over recent-, frequent-, and anchor-seeking actions, together with an abstention option.
Finally, we develop a behavior-conditioned activity realization module that generates the next timestamped activity at a concrete target POI. A dynamic action--POI graph propagates action probabilities to behaviorally compatible candidates, providing permutation-equivariant grounding across region-specific POI vocabularies. A consistency-guided product-of-experts (PoE) then integrates the grounded behavioral preference with prefix consistency and identity-free transition evidence to generate the final activity.

We extensively evaluate \m on large-scale real-world HATs from ten U.S. cities. Our evaluation examines downstream utility, generation fidelity, component contributions, source-region sensitivity, and computational efficiency.
Across the three main target regions, \m achieves $4.5$--$6.4\times$ the average normalized utility of the strongest baseline and attains the best aggregate fidelity, reducing average JSD by $15.6\%$--$40.8\%$.
Ablation studies show that the cross-region behavioral cloning module improves downstream utility, while the consistency intents help preserve spatial fidelity; combining both with identity-free transition evidence through PoE realization achieves the highest average utility while retaining strong fidelity.
Source-region experiments further show that transferable behavioral structures interact differently with each target context: concentrated source behaviors generally favor downstream utility, whereas more diverse mixtures improve fidelity.
Finally, \m reaches around $3\times$ throughput of the fastest neural baseline while using only $6.74$ GB of peak GPU memory, compared with $39.46$--$48.52$ GB for the neural baselines.

Our main contributions are summarized as follows:
\begin{itemize}
    \item Conceptually, we focus on zero-shot HAT generation, which generates synthetic HATs for a target region using authentic source-region HATs and publicly available context, without requiring any target-region HATs.

       \item Technically, we propose \m, a behavior-conditioned zero-shot HAT generation framework with three novel components: a multidimensional consistency-aware intent extractor that extracts temporal, semantic, and relative spatial consistency intents from generated prefixes; a cross-region behavioral cloning module that learns approximately region-invariant actions for revisiting recent, frequent, or anchor POIs; and a behavior-conditioned activity realization module that generates the next timestamped activity at a concrete target POI.
       
    \item Empirically, we evaluate \m on real HAT data from ten cities. Experiments show that \m achieves $4.5$--$6.4\times$ the utility of the strongest baseline and reduces average JSD by $15.6\%$--$40.8\%$. Further analyses validate the model design, characterize source-region sensitivity, and demonstrate efficient generation.

\end{itemize}

% Map2Traj generates coordinate trajectories from street maps~\citep{map2traj2024}, while MobTA transfers timetable-induced mobility changes through region-relative grid tokens~\citep{mobta2026}.
% Although these methods avoid target-region trajectories, their outputs are coordinate paths or grid sequences rather than semantically rich HATs.

%% Behavioral patterns learned from source regions must generalize to a high-dimensional target region context shaped by distinct POIs, spatial layouts, category supply, popularity, accessibility, and coarse spatial flows, without individual target HATs revealing how these environmental components interact.
%retains $53.1\%$--$75.1\%$ of the utility obtained with real target-region HATs, 
\section{Preliminaries}
\subsection{Human Activity Trace}
Let $\mathcal{C}$ denote a set of regions (e.g., cities).
For each region $c \in \mathcal{C}$, let $\mathcal{P}_c$ be its POI set.
A POI $p \in \mathcal{P}_c$ is associated with attributes $x_p$, including geographic coordinates (latitude and longitude), POI activity category, and visit popularity.
A human activity trace (HAT) in region $c$ is a timestamped POI sequence of a user
\begin{equation}
    h = \big((p_1,t_1), (p_2,t_2), \ldots, (p_L,t_L)\big),
\end{equation}
where $p_i \in \mathcal{P}_c$ is the visited POI at time $t_i$, and $L$ is the trace length.
We use $\mathcal{H}_c = \{h_n\}_{n=1}^{N_c}$ to denote the set of observed HATs in region $c$.
Since POI sets are region-specific, $\mathcal{P}_c$ and $\mathcal{P}_{c'}$ generally have different identities and sizes when $c \neq c'$.

\subsection{Zero-Shot HAT Generation}
We consider a leave-region-out setting with source regions $\mathcal{C}_s$ and a target region $c_t \notin \mathcal{C}_s$.
For each source region $c \in \mathcal{C}_s$, the model can access both observed HATs $\mathcal{H}_c$ and publicly available region context $\mathcal{X}_c = \{\mathcal{P}_c, F_c\}$, where $F_c$ denotes some aggregated information such as a coarse cell-level origin-destination (OD)-flow prior that describes aggregate movement tendency between spatial cells in region $c$.
For the target region $c_t$, the model can access only target context $\mathcal{X}_{c_t}$ and cannot use any target-region HATs during training or generation.
The training input is therefore $\mathcal{D}_s = \{(\mathcal{H}_c, \mathcal{X}_c): c \in \mathcal{C}_s\}$,
and the inference input is $\mathcal{X}_{c_t}$.
The goal is to generate a synthetic target-region HAT set
\begin{equation}
    \widehat{\mathcal{H}}_{c_t} = \{\hat{h}_m\}_{m=1}^{M}, \qquad \hat{h}_m \in (\mathcal{P}_{c_t} \times \mathcal{T})^{*},
\end{equation}
where $\mathcal{T}$ denotes the timestamp space and $M$ is the number of generated traces.
\section{Methodology}
\begin{figure*}[t]
    \centering
    \includegraphics[width=\textwidth]{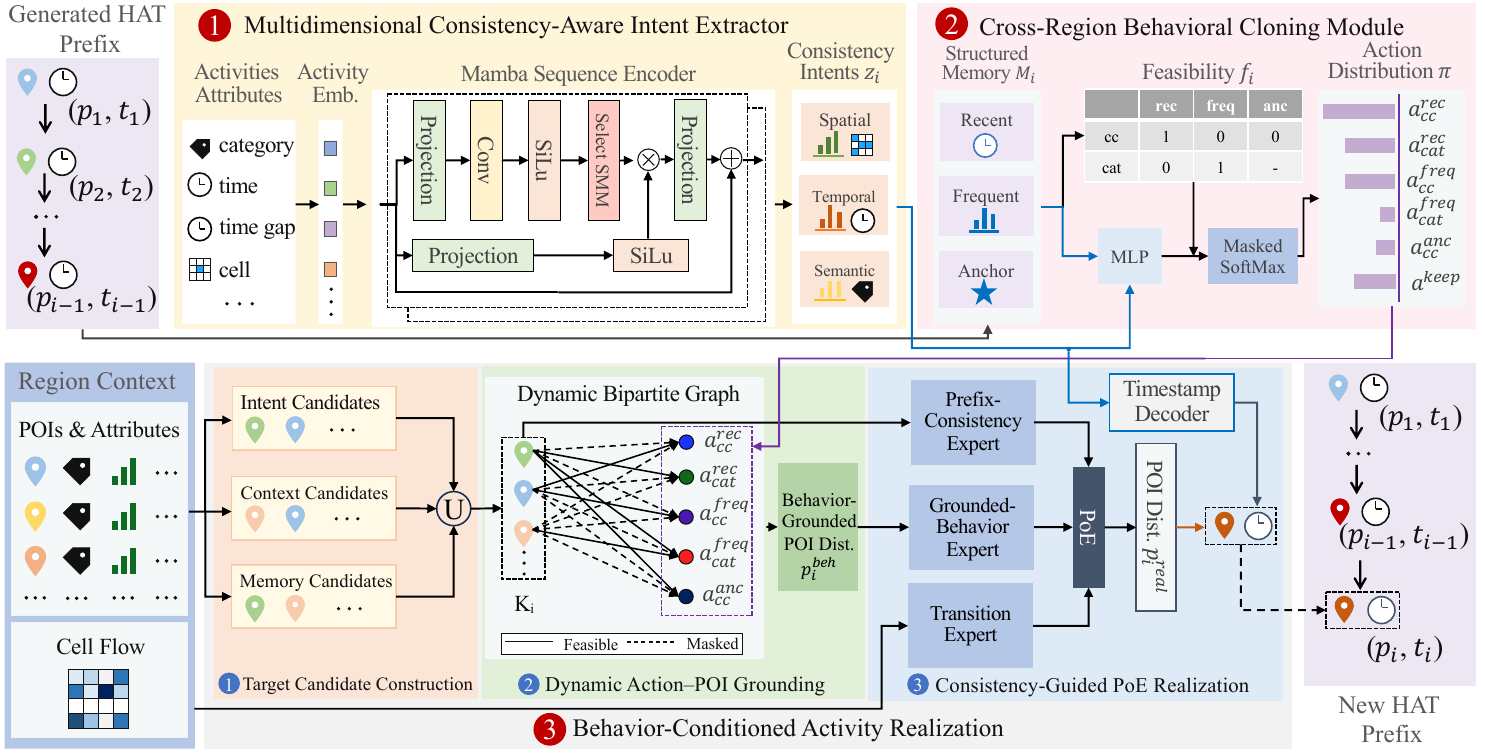}
    \caption{The proposed \m framework.}
    \label{fig:framework}
\end{figure*}

Fig.~\ref{fig:framework} shows the overall framework of \m, which includes three key modules: 
a multidimensional consistency-aware intent extractor that produces temporal, semantic, and relative spatial consistency intents;
a cross-region behavioral cloning policy that generates approximately region-invariant behavioral action distributions;
and a behavior-conditioned activity realization module that finalizes the next activity. \m generates synthetic HATs in an autoregressive way, where each next activity is generated based on the previously generated prefix.

\subsection{Multidimensional Consistency-Aware Intent Extractor}
A realistic next activity should be temporally, semantically, and spatially consistent with its prefix.
To better model these constraints, we encode the prefix into a multidimensional representation that summarizes the current HAT state and predicts next-activity consistency intent.
To avoid dependence on a region-specific POI vocabulary, we use \textit{region-agnostic features} of each POI and its surrounding context.
Let $\hat{h}_{<i}=((\hat{p}_1,\hat{t}_1),\ldots,(\hat{p}_{i-1},\hat{t}_{i-1}))$ be the generated prefix at step $i$.
For activity $j<i$, we construct
\begin{equation}
\begin{split}
    e_j=\phi_e\!\bigl([&E_{\mathrm{poi}}(\hat p_j),
    E_{\mathrm{cell}}(\hat p_j),E_{\mathrm{cat}}(\hat p_j),\\
    &E_{\mathrm{time}}(\hat t_j),
    E_{\mathrm{gap}}(\hat t_j,\hat t_{j-1})]\bigr).
\end{split}
\end{equation}
Here $E_{\mathrm{poi}}$ extracts identifier-free POI attributes such as relative location, popularity, and accessibility; $E_{\mathrm{cell}}$ encodes the surrounding cell through its category supply and coarse flow; and $E_{\mathrm{cat}}$ embeds the POI category;
$E_{\mathrm{time}}$ encodes the absolute timestamp; and $E_{\mathrm{gap}}$ encodes the elapsed time since the previous activity. $\phi_e$ projects their concatenation into a shared event space.
We adopt a Mamba-based state-space encoder~\cite{gu2024mamba} to model the event sequence, as its selective state-space formulation captures long-range dependencies across multidimensional event representations with linear complexity in sequence length:
\begin{equation}
\begin{split}
    (m_1,\ldots,m_{i-1})
    &=\operatorname{Mamba}(e_1,\ldots,e_{i-1}),\\
    h_i&=m_{i-1}.
\end{split}
\end{equation}
Here $h_i$ summarizes the current HAT state.
We decouple the next-activity consistency intent into temporal, semantic, and relative spatial components.
The temporal heads predict an absolute-time distribution $P_i^{\mathrm{time}}=\operatorname{softmax}(W_t h_i)$
and a time-gap distribution $P_i^{\mathrm{gap}}=\operatorname{softmax}(W_{\Delta}h_i)$,
while the semantic head predicts a category distribution $P_i^{\mathrm{cat}}=\operatorname{softmax}(W_q h_i)$.
For a target candidate $p$, the spatial head evaluates relative compatibility rather than predicting a globally indexed cell,
\begin{equation}
    s_i^{\mathrm{sp}}(p)=
    \phi_{\mathrm{sp}}\!\left(h_i,x_p,
    \rho(p,\hat p_{i-1})\right),
\end{equation}
where $\rho$ contains candidate-relative distance, same-cell, accessibility, and coarse-flow relations.
We denote the resulting consistency intent by
\begin{equation}
    z_i=\left(P_i^{\mathrm{time}},P_i^{\mathrm{gap}},
    P_i^{\mathrm{cat}},s_i^{\mathrm{sp}}(\cdot)\right).
\end{equation}
Thus, $z_i$ specifies when the next activity is likely to occur, which semantic role it should serve, and which relative spatial relations are consistent with the prefix.

\subsection{Cross-Region Behavioral Cloning}
We found that many behavioral structures, such as returning to recent, frequent, or regular locations, are transferable across regions.
We capture them through behavioral cloning over a factorized, region-invariant action space.
Each memory-seeking action is defined by a behavior operator $b$ and a retrieval scope $\kappa$:
\begin{equation}
\begin{split}
    \mathcal{B}&=\{\mathrm{rec},\mathrm{freq},\mathrm{anc}\},\qquad
    \mathcal{S}=\{\mathrm{cc},\mathrm{cat}\},\\
    \Omega&=(\mathcal{B}\times\mathcal{S})
    \setminus\{(\mathrm{anc},\mathrm{cat})\},\\
    \mathcal{A}&=\{a^{\mathrm{keep}}\}
    \cup\{a_{\kappa}^{b}:(b,\kappa)\in\Omega\}.
\end{split}
\end{equation}
Here $b$ specifies whether to retrieve a recent ($\mathrm{rec}$), frequent ($\mathrm{freq}$), or anchor ($\mathrm{anc}$) POI in the prefix memory, while $\kappa$ specifies a cell--category ($\mathrm{cc}$) or category-only ($\mathrm{cat}$) scope.
The special action $a^{\mathrm{keep}}$ abstains from memory retrieval and lets the realization module rely on consistency and target context.
At step $i$, we construct a structured prefix memory
$\mathcal{M}_i=(\mathcal{M}_i^{\mathrm{rec}},\mathcal{M}_i^{\mathrm{freq}},\mathcal{M}_i^{\mathrm{anc}})$, which stores visit order, visit counts, and first-observed anchors under category and cell--category keys of the prefix HAT.
We decode the soft intent $z_i$ into a working scope
$\tilde z_i=(\tilde c_i,\tilde q_i)$.
Specifically,
$\tilde q_i=\operatorname*{argmax}_{q}P_i^{\mathrm{cat}}(q)$
is the most likely category, while
$\tilde c_i=\operatorname*{argmax}_{c}
\sum_{p:c_p=c}\exp\!\left(s_i^{\mathrm{sp}}(p)\right)$
is the cell that best matches the spatial signal.
The corresponding scope queries are
\begin{equation}
\begin{aligned}
\mathcal{Q}_{\mathrm{cc}}(\mathcal{M}_i,z_i)
&=\{\hat p_j:j<i,\,
(c_{\hat p_j},q_{\hat p_j})=\tilde z_i\},\\
\mathcal{Q}_{\mathrm{cat}}(\mathcal{M}_i,z_i)
&=\{\hat p_j:j<i,\,
q_{\hat p_j}=\tilde q_i\}.
\end{aligned}
\end{equation}
All non-keep actions instantiate the same query--selection operator:
\begin{equation}
g_{a_{\kappa}^{b}}(\mathcal{M}_i,z_i)
=\operatorname{Select}_{b}
\bigl(\mathcal{Q}_{\kappa}(\mathcal{M}_i,z_i)\bigr).
\end{equation}
The operators $\operatorname{Select}_{\mathrm{rec}}$,
$\operatorname{Select}_{\mathrm{freq}}$, and
$\operatorname{Select}_{\mathrm{anc}}$ return the most recent,
most frequently visited, and first-observed POI, respectively.
An action is feasible when its query is nonempty; $f_i(a)$ records this condition, with $f_i(a^{\mathrm{keep}})=1$.
The policy combines the encoded HAT state $h_i$, consistency intent $\eta(z_i)$, memory statistics $\mu(\mathcal{M}_i)$, and action feasibility $f_i$:
\begin{equation}
\begin{split}
    u_i&=[h_i\,\|\,\eta(z_i)\,\|\,\mu(\mathcal{M}_i)\,\|\,f_i],\\
    \pi_{\theta}(a_i\mid u_i)
    &=\operatorname{MaskedSoftmax}
    (\operatorname{MLP}_{\theta}(u_i);f_i),
\end{split}
\end{equation}
where $\eta$ encodes the consistency distributions and $\mu$ summarizes the structured memory.
The masked softmax assigns zero probability to infeasible actions before normalization.
We derive cloning targets by replaying each source-region trace.
The policy is optimized with
\begin{equation}
    \mathcal{L}_{\mathrm{BC}}=-
    \sum_{c\in\mathcal{C}_s}\sum_{h\in\mathcal{H}_c}
    \sum_{i=1}^{|h|}
    \log \pi_{\theta}(a_i^{\star}\mid u_i).
\end{equation}
Consequently, we obtain a probability distribution over the candidate actions.

\subsection{Behavior-Conditioned Activity Realization}
Given the consistency intent $z_i$, action distribution $\pi_\theta$, and target-region context $\mathcal{X}_{c_t}$, this module constructs candidate target POIs, grounds the action distribution over them through a dynamic bipartite graph, and integrates the grounded behavioral preferences with consistency and transition evidence to generate the next activity.

\paragraph{Target Candidate Construction.}
We build a variable-size set
$\mathcal{K}_i=\mathcal{K}_i^{\mathrm{con}}\cup\mathcal{K}_i^{\mathrm{ctx}}\cup\mathcal{K}_i^{\mathrm{mem}}\subseteq\mathcal{P}_{c_t}$.
The three pools contain POIs supported by the semantic--spatial intent, public target context, and feasible prefix-memory queries, respectively (see Appendix for retrieval and pruning details).
For $p\in\mathcal{K}_i$, a consistency score combines the semantic--spatial intent with static target context:
\begin{equation}
\begin{split}
    S_i^{\mathrm{con}}(p)=&\ s_i^{\mathrm{sp}}(p)
    +\lambda_q\log P_i^{\mathrm{cat}}(q_p)\\
    &+\lambda_x\phi_x(x_p,\mathcal{X}_{c_t}).
\end{split}
\end{equation}
Here $q_p$ is the category of $p$, while $\phi_x$ encodes its popularity, category supply, and accessibility in the target region.
Applying normalization yields
\begin{equation}
    P_i^{\mathrm{con}}(p)=
    \frac{\exp S_i^{\mathrm{con}}(p)}
    {\sum_{p'\in\mathcal{K}_i}\exp S_i^{\mathrm{con}}(p')}.
\end{equation}

\paragraph{Dynamic Action--POI Grounding.}
We introduce dynamic action--POI grounding to map each action and its probability $\pi_\theta$ to compatible POIs.
At step $i$, we construct a dynamic bipartite graph
$\mathcal{G}_i=(\mathcal{A}_i^{+},\mathcal{K}_i,\mathcal{E}_i)$,
where $\mathcal{A}_i^{+}$ contains feasible non-keep action nodes,
$\mathcal{K}_i$ contains target-POI nodes, and
$\mathcal{E}_i=\{(a,p):\mu_i(a,p)=1\}$ contains admissible
action--POI edges.
The mask $\mu_i(a,p)$ enforces the retrieval scope $\kappa$ and the
memory evidence required by behavior $b$.
For each admissible edge, we compute transport mass $T_i(a,p)$ from action $a$ to POI $p$ as:
\begin{equation}
\begin{aligned}
E_i(a,p)
&=u_a^{\top}W_g\psi_i(p),\\
G_i(p\mid a)
&=\frac{\mu_i(a,p)\exp(E_i(a,p)/\tau)}
{\sum_{p'\in\mathcal{K}_i}
 \mu_i(a,p')\exp(E_i(a,p')/\tau)},\\
T_i(a,p)
&=\pi_\theta(a\mid u_i)G_i(p\mid a),
\end{aligned}
\end{equation}
where $u_a$ is an action embedding and $\psi_i(p)$ contains recency,
frequency, anchor membership, category agreement, relative spatial
consistency, coarse-flow, popularity, supply, and accessibility.
The resulting transport mass satisfies
$T_i(a,p)=0$ for inadmissible edges and
$\sum_{p}T_i(a,p)=\pi_\theta(a\mid u_i)$.
Thus, the graph preserves the learned behavioral preference while
softly distributing each action over multiple compatible target POIs.
The transported action mass forms the behavior-grounded POI
distribution:
\begin{equation}
P_i^{\mathrm{beh}}(p)
=
\pi_\theta(a^{\mathrm{keep}}\mid u_i)P_i^{\mathrm{con}}(p)
+
\sum_{a\in\mathcal{A}_i^{+}}T_i(a,p).
\end{equation}
Because the edge scorer is shared across POIs and uses no POI
identifiers, this transport is \textit{permutation-equivariant} over variable
target POI sets and requires no cross-region POI alignment.

\begin{table*}[tp]
\centering
\small
\caption{Downstream utility ratios for three representative target cities across four tasks: DeepMove (DM), STAN, GetNext (GN), and STARHIT (SH), and their average (Avg.). Higher is better; \best{red} and \textcolor{blue}{blue} indicate the best and second-best results in each column, respectively.}
\label{tab:utility}
\resizebox{\textwidth}{!}{%
\begin{tabular}{l|c c c c c|c c c c c|c c c c c}
\hline
& \multicolumn{5}{c|}{Atlanta} & \multicolumn{5}{c|}{Houston} & \multicolumn{5}{c}{Seattle} \\
\cline{2-16}
Method & DM & STAN & GN & SH & Avg. & DM & STAN & GN & SH & Avg. & DM & STAN & GN & SH & Avg. \\
\hline
Markov & 0.1095 & 0.0684 & \second{0.0684} & 0.0988 & 0.0863 & 0.2079 & 0.0811 & 0.0925 & 0.1501 & 0.1329 & 0.0993 & 0.0671 & 0.0739 & 0.0944 & 0.0837 \\
TimeGeo & \second{0.1364} & \second{0.0791} & 0.0646 & \second{0.1128} & \second{0.0982} & \second{0.2663} & \second{0.0965} & \second{0.1093} & \second{0.1943} & \second{0.1666} & \second{0.1187} & 0.0585 & \second{0.0825} & \second{0.1113} & \second{0.0927} \\
LSTM & 0.0677 & 0.0452 & 0.0506 & 0.0409 & 0.0511 & 0.0988 & 0.0734 & 0.0388 & 0.0559 & 0.0667 & 0.0587 & 0.0334 & 0.0439 & 0.0387 & 0.0437 \\
MobFormer & 0.0593 & 0.0429 & 0.0444 & 0.0495 & 0.0490 & 0.0759 & 0.0356 & 0.0297 & 0.0636 & 0.0512 & 0.0355 & 0.0278 & 0.0293 & 0.0296 & 0.0306 \\
COLA & 0.0784 & 0.0493 & 0.0579 & 0.0582 & 0.0610 & 0.1071 & 0.0412 & 0.0724 & 0.0691 & 0.0725 & 0.0636 & 0.0588 & 0.0373 & 0.0553 & 0.0538 \\
MobTA & 0.0689 & 0.0421 & 0.0580 & 0.0678 & 0.0592 & 0.1100 & 0.0465 & 0.0696 & 0.0928 & 0.0797 & 0.0438 & \second{0.1295} & 0.0263 & 0.0601 & 0.0649 \\
ActivityEditor & 0.0594 & 0.0659 & 0.0321 & 0.0628 & 0.0550 & 0.1064 & 0.0284 & 0.0207 & 0.0517 & 0.0518 & 0.0435 & 0.0311 & 0.0247 & 0.0535 & 0.0382 \\
\textsc{ZeroHAT} & \best{0.5225} & \best{0.5383} & \best{0.5334} & \best{0.5308} & \best{0.5313} & \best{0.7464} & \best{0.7482} & \best{0.7976} & \best{0.7128} & \best{0.7512} & \best{0.5651} & \best{0.5664} & \best{0.5869} & \best{0.6371} & \best{0.5889} \\
\hline
\end{tabular}%
}
\end{table*}
\begin{table*}[tp]
\centering
\small
\caption{Generation fidelity against held-out real HATs on three representative target cities. Lower is better.}
\label{tab:fidelity}
\resizebox{\textwidth}{!}{%
\begin{tabular}{l|c c c c|c c c c|c c c c}
\hline
& \multicolumn{4}{c|}{Atlanta} & \multicolumn{4}{c|}{Houston} & \multicolumn{4}{c}{Seattle} \\
\cline{2-13}
Method & Spatial & Temporal & Semantic & Avg. & Spatial & Temporal & Semantic & Avg. & Spatial & Temporal & Semantic & Avg. \\
\hline
Markov & 0.4086 & 0.0035 & 0.1851 & \second{0.1991} & 0.4867 & \second{0.0032} & \second{0.1809} & \second{0.2236} & 0.4469 & 0.0027 & 0.1512 & \second{0.2003} \\
TimeGeo & \second{0.3971} & 0.1817 & \second{0.1732} & 0.2507 & \second{0.4687} & 0.1816 & \best{0.1732} & 0.2745 & \second{0.4459} & 0.1821 & \second{0.1424} & 0.2568 \\
LSTM & 0.4156 & 0.0060 & 0.1882 & 0.2032 & 0.4876 & 0.0032 & 0.1865 & 0.2258 & 0.4656 & 0.0039 & 0.1528 & 0.2074 \\
MobFormer & 0.4203 & \second{0.0032} & 0.1986 & 0.2074 & 0.4923 & 0.0035 & 0.1909 & 0.2289 & 0.4689 & \second{0.0025} & 0.1634 & 0.2116 \\
COLA & 0.4127 & 0.0041 & 0.1898 & 0.2022 & 0.4847 & 0.0032 & 0.1857 & 0.2245 & 0.4703 & 0.0028 & 0.1557 & 0.2096 \\
MobTA & 0.4147 & 0.0068 & 0.1952 & 0.2056 & 0.4810 & 0.0034 & 0.1947 & 0.2264 & 0.4711 & 0.0047 & 0.1597 & 0.2118 \\
ActivityEditor & 0.4171 & 0.1795 & 0.1963 & 0.2643 & 0.4924 & 0.1799 & 0.1962 & 0.2895 & 0.4758 & 0.1785 & 0.1628 & 0.2723 \\
\textsc{ZeroHAT} & \best{0.2259} & \best{0.0007} & \best{0.1404} & \best{0.1224} & \best{0.3763} & \best{0.0008} & 0.1891 & \best{0.1888} & \best{0.2359} & \best{0.0002} & \best{0.1193} & \best{0.1185} \\
\hline
\end{tabular}
}
\end{table*}

\paragraph{Consistency-Guided PoE Realization.}
A POI that realizes a habitual action may still be inconsistent with the
evolving trace.
To reconcile these signals, we introduce a consistency-guided
product-of-experts (PoE) realization that integrates three experts: a prefix-consistency
expert $\exp(S_i^{\mathrm{con}}(p))$, a grounded-behavior expert $\epsilon+P_i^{\mathrm{beh}}(p)$, and an identity-free transition expert $\exp(S_i^{\mathrm{trans}}(p))$.
The transition expert complements them with source-region category dynamics
and public coarse flow:
\begin{equation}
\begin{split}
S_i^{\mathrm{trans}}(p)=&\
\log\!\left(\epsilon+
P_{\mathcal{C}_s}^{\mathrm{cat}}
(q_p\mid q_{\hat p_{i-1}})\right)\\
&+\beta_f\log\!\left(
\epsilon+F_{c_t}(c_{\hat p_{i-1}},c_p)\right),
\end{split}
\end{equation}
where $P_{\mathcal{C}_s}^{\mathrm{cat}}$ is estimated from source-region
HATs and $F_{c_t}$ is the target coarse-flow prior;
$q_{\hat p_{i-1}}$ and $c_{\hat p_{i-1}}$ denote the category and region of
the preceding activity.
Here $c_p$ denotes the region of $p$, $\epsilon>0$ prevents zero probabilities.
The three experts form an unnormalized PoE score $\Phi_i$, which is normalized into $P_i^{\mathrm{real}}$:
\begin{equation}
\begin{split}
\Phi_i(p)=&\
\exp\!\left(S_i^{\mathrm{con}}(p)\right)
\left(\epsilon+P_i^{\mathrm{beh}}(p)\right)^{\lambda_b}\\
&\times\exp\!\left(\lambda_T S_i^{\mathrm{trans}}(p)\right),\\
P_i^{\mathrm{real}}(p)=&\
\frac{\Phi_i(p)}
{\sum_{p'\in\mathcal{K}_i}\Phi_i(p')},
\end{split}
\end{equation}
where $\beta_f$, $\lambda_b$, and $\lambda_T$ weight coarse-flow,
behavior, and transition evidence, respectively.
Unlike mixture-of-experts (MoE), PoE assigns high probability only to POIs jointly supported by the behavioral, sequential, and transition experts, which ensures the multidimensional consistency of the realized POIs.

\subsection{Source-Only Training}
We train \m under teacher forcing on each source HAT, using
the observed prefix $h_{<i}$ and including its next POI $p_i$ in
$\mathcal{K}_i$ during training.
The source-only objective is
\begin{equation}
\begin{split}
\mathcal{L}={}&\sum_{c\in\mathcal{C}_s}\sum_{h\in\mathcal{H}_c}
\sum_{i=1}^{|h|}\Bigl(\ell_i^{\mathrm{poi}}
+\lambda_t\ell_i^{\mathrm{time}}+\lambda_{\Delta}\ell_i^{\mathrm{gap}}\\
&\qquad+\lambda_q\ell_i^{\mathrm{cat}}+\lambda_s\ell_i^{\mathrm{sp}}
+\lambda_{\mathrm{BC}}\ell_i^{\mathrm{BC}}\Bigr).
\end{split}
\end{equation}
Here $\ell_i^{\mathrm{poi}}=-\log P_i^{\mathrm{real}}(p_i)$,
$\ell_i^{\mathrm{BC}}=-\log\pi_\theta(a_i^\star\mid u_i)$; the time, gap,
and category terms are cross-entropies for the observed activity.
The spatial loss normalizes relative compatibility over the source candidate set:
\begin{equation}
\ell_i^{\mathrm{sp}}=-\log
\frac{\exp s_i^{\mathrm{sp}}(p_i)}
{\sum_{p\in\mathcal{K}_i}\exp s_i^{\mathrm{sp}}(p)}.
\end{equation}
Complete cloning-label construction and optimization details are provided in
Appendix.
% ~\ref{app:training_details}.

\section{Evaluation}
\subsection{Evaluation Setup}
\subsubsection{Datasets.}
We evaluate \m on real-world HATs from ten U.S.
regions provided by Dewey platform~\citep{deweydata2026}.
The main setting uses Boston and Chicago as source regions and reports results
on three representative target regions: Atlanta, Houston, and Seattle; the
other five held-out regions are used for robustness diagnostics.
Detailed data statistics and preprocessing are under \emph{Dataset Details} in the Appendix.

\subsubsection{Baselines.}
We compare \m against seven state-of-the-art baselines in three groups: (1)
statistical and physics-based methods including Markov and TimeGeo~\citep{timegeo2016};
(2) sequential models including LSTM and MobFormer; and (3) zero-shot generation models includingCOLA~\citep{cola2024}, MobTA~\citep{mobta2026}, and
ActivityEditor~\citep{activityeditor2026}. Details of these methods are in \emph{Baseline Details} in the Appendix.

\subsubsection{Metrics.}
We comprehensively evaluate downstream utility, generation fidelity, and efficiency.
We evaluate utility on the downstream POI recommendation task. We utilize the Recall@$k$ ratio between a model trained on
synthetic target HATs and the same model trained on real target HATs, evaluated
on an identical held-out real test set. 
We evalaute it on four models: DeepMove \citep{deepmove2018}, STAN \citep{stan2021}, GetNext \citep{getnext2022}, and STARHIT \citep{starhit2022}.
We evaluate fidelity using Jensen--Shannon
divergence (JSD) over spatial, temporal, and semantic distributions; lower is
better. Efficiency is measured by generation throughput and peak GPU memory on one NVIDIA B200 GPU.
Details appear under \emph{Metric Details} in the Appendix.

\begin{figure*}[!tp]
\centering
\includegraphics[width=\textwidth]{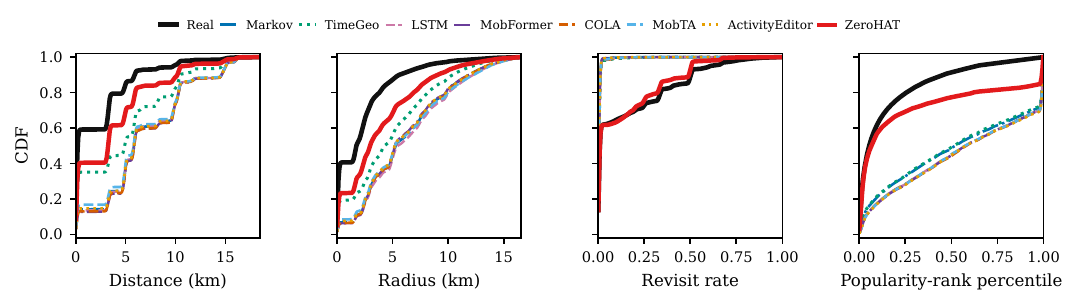} \vspace{-15pt}
\caption{Travel distance, trace radius, revisit rate, and visited-POI popularity-rank percentile distribution of generated HATs in Seattle.}
\label{fig:raw_cdf_panel}
% \includegraphics[width=\textwidth]{figures/raw_hat_nature/fig_raw_transition_matrix_seattle.pdf}
% \caption{Coarse-category transition density matrices on Seattle. Panels show (a) Real, (b) MobTA, (c) ActivityEditor, and (d) \m.}
% \label{fig:raw_transition_matrix}
\end{figure*}

% \vspace{-3pt}
\subsection{RQ1: Downstream Application Utility}
Table~\ref{tab:utility} shows that \m consistently delivers large utility gains over the baselines across the three target cities. On average, \m achieves $5.4\times$, $4.5\times$, and $6.4\times$ the utility of the strongest baseline on Atlanta, Houston, and Seattle, respectively. These improvements hold across all four downstream models. DeepMove and STAN emphasize sequential history, whereas GetNext and STARHIT adopt different spatio-temporal ranking mechanisms, demonstrating the robustness of \m across diverse downstream models. The strongest baseline remains below $0.17$ average utility in all three cities despite using the same target POI context. These consistent gains suggest that \m preserves transferable behavioral structures that remain useful across heterogeneous tasks, beyond target grounding alone.

\begin{table}[t]
\centering
\caption{Ablation results averaged over the three target regions.}
\label{tab:ablation}
\resizebox{\columnwidth}{!}{%
\begin{tabular}{@{}l c c c c c@{}}
\hline
Variant & DM & ST & GN & SH & Avg. \\
\hline
Consistency Only & 0.5326 & 0.5176 & 0.5140 & 0.4987 & 0.5157 \\
Direct Behavior & \best{0.5623} & \best{0.5589} & \second{0.5902} & 0.5575 & \second{0.5673} \\
Weighted Fusion & \second{0.5596} & 0.5414 & 0.5818 & 0.5801 & 0.5657 \\
Adaptive Fusion & 0.5305 & 0.5419 & 0.5628 & \second{0.5860} & 0.5553 \\
w/o Transition & 0.5399 & 0.5538 & 0.5751 & 0.5857 & 0.5636 \\
\m & 0.5573 & \second{0.5547} & \best{0.5939} & \best{0.5916} & \best{0.5744} \\
\hline
\end{tabular}%
}
\end{table}

\subsection{RQ2: Generation Fidelity}
As shown in Table~\ref{tab:fidelity}, \m achieves the best average generation fidelity among all compared methods across the three target cities.
Against the strongest competing average-fidelity baseline, \m reduces average JSD by $38.5\%$ on Atlanta, $15.6\%$ on Houston, and $40.8\%$ on Seattle.
The most substantial absolute improvements occur in spatial fidelity: POI-distribution JSD drops by $43.1\%$, $19.7\%$, and $47.1\%$ relative to the strongest spatial baseline on Atlanta, Houston, and Seattle.
\m also obtains the best temporal JSD on all three cities.
% and the best semantic JSD on Atlanta and Seattle.
Figures~\ref{fig:raw_cdf_panel} provide visual evidence.
The CDFs show closer alignment with real traces in travel distance, trace radius, revisit rate, and visited-POI popularity rank.
Additional transition matrices in Appendix Figures~\ref{fig:appendix_spatial_density_atlanta}, \ref{fig:appendix_spatial_density_houston}, and~\ref{fig:appendix_spatial_density_seattle} also show that \m preserves multiple high-density within-category and cross-category transition regions rather than collapsing probability mass into a single coarse activity block. 

% Additional coarse-category POI-density maps in Appendix Figures~\ref{fig:appendix_spatial_density_atlanta}, \ref{fig:appendix_spatial_density_houston}, and~\ref{fig:appendix_spatial_density_seattle} show concentration patterns similar to those of real traces around major target-city POI clusters.

\subsection{RQ3: Ablation Study}
We examine the contributions of behavioral cloning, consistency guidance, transition modeling, and expert fusion to \m. All variants are evaluated using the same four downstream models as in RQ1. \emph{Consistency Only} removes behavioral cloning and action--POI grounding and realizes activities using consistency evidence alone. \emph{Direct Behavior} directly uses behavior-grounded outcomes without the final consistency-guided PoE. \emph{Weighted Fusion} replaces the multiplicative PoE with an additive weighted combination of the same evidence, while \emph{Adaptive Fusion} makes the fusion weights dependent on the target context. \emph{w/o Transition} sets $\lambda_T=0$ to remove the identity-free transition expert.

% We examine how behavioral cloning, consistency guidance, transition modeling, and expert fusion contribute to \m.
% All variants are evaluated with the same four downstream predictors used in RQ1.
% \emph{Consistency Only} removes behavioral cloning and action--POI grounding, and realizes activities from consistency evidence alone.
% \emph{Direct Behavior} directly uses behavior-grounded outcomes without the final consistency-guided PoE.
% \emph{Weighted Fusion} replaces the multiplicative PoE with an additive weighted combination of the same evidence, while \emph{Adaptive Fusion} makes the fusion weights target-context dependent.
% \emph{w/o Transition} sets $\lambda_T=0$ to remove the identity-free transition expert.
% \m retains all components with fixed, source-validated PoE weights.

Table~\ref{tab:ablation} demonstrates the complementary roles of consistency guidance and transferable behavior. \emph{Consistency Only} yields the lowest average utility ($0.5157$), whereas \emph{Direct Behavior} increases it to $0.5673$, confirming the value of behavioral cloning and action--POI grounding. The full model achieves the highest average utility ($0.5744$), including the best scores on GetNext ($0.5939$) and STARHIT ($0.5916$). \emph{Weighted Fusion} performs worse on average ($0.5657$), supporting the PoE design, which favors POIs jointly supported by consistency, behavior, and transition evidence rather than combining their scores additively. \emph{Adaptive Fusion} further reduces average utility to $0.5553$, suggesting that target-context-dependent weighting is less robust than fixed, source-validated fusion in the zero-shot setting. Finally, removing the transition expert lowers average utility to $0.5636$, demonstrating its contribution to behavior-grounded activity realization. Fidelity remains comparable across all variants, with only minor differences in the grouped JSD metrics.

\vspace{-6pt}
\begin{figure}[!h]
\centering
\begin{subfigure}[t]{0.48\columnwidth}
\centering
\includegraphics[width=\linewidth]{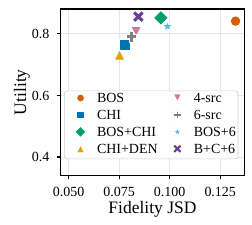}
\caption{Houston}
\end{subfigure}
\hfill
\begin{subfigure}[t]{0.48\columnwidth}
\centering
\includegraphics[width=\linewidth]{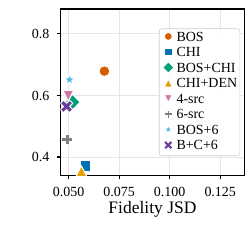}
\caption{Seattle}
\end{subfigure}
\caption{Source-composition sensitivity in zero-shot transfer.}
\label{fig:rq4_source_sensitivity_top}
\end{figure}

\vspace{-12pt}
\subsection{RQ4: Source-Composition Sensitivity}

We investigate how different combinations of real HATs from diverse source cities affect performance across the three target cities.
The source pool comprises Boston (BOS), Chicago (CHI), Denver (DEN), Phoenix (PHX), Los Angeles (LAX), and New York (NYC).
We compare BOS, CHI, BOS+CHI, CHI+DEN, an equal-share four-source pool (BOS, CHI, DEN, and PHX), and an equal-share six-source pool.
Two anchor--background mixtures, BOS+6 and BOS+CHI+6, assign 50\% and 25\% of events to their anchor components, respectively.
Diagnostic utility averages the DeepMove and GetNext Recall@$10$ utility ratios, while fidelity is mean JSD; complete results appear in Appendix.
Figure~\ref{fig:rq4_source_sensitivity_top} shows that source composition shifts the utility--fidelity operating point differently for Houston and Seattle.

% \begin{figure}[!tp]
% \centering
% \begin{subfigure}[t]{0.48\columnwidth}
% \centering
% \includegraphics[width=\linewidth]{figures/rq4/fig_rq4_sub_a_houston.pdf}
% \caption{Houston}
% \end{subfigure}
% \hfill
% \begin{subfigure}[t]{0.48\columnwidth}
% \centering
% \includegraphics[width=\linewidth]{figures/rq4/fig_rq4_sub_b_seattle.pdf}
% \caption{Seattle}
% \end{subfigure}
% \vspace{2pt}
% \begin{subfigure}[t]{0.48\columnwidth}
% \centering
% \includegraphics[width=\linewidth]{figures/rq4/fig_rq4_sub_c_city_behavior.pdf}
% \caption{City behavior}
% \end{subfigure}
% \hfill
% \begin{subfigure}[t]{0.48\columnwidth}
% \centering
% \includegraphics[width=\linewidth]{figures/rq4/fig_rq4_sub_d_mechanism.pdf}
% \caption{Behavior concentration}
% \end{subfigure}
% \caption{Source-composition sensitivity in zero-shot transfer. Panels (a,b) show target-specific utility--fidelity operating points; (c) summarizes city behavior priors; and (d) relates generated behavior concentration to utility and fidelity across source settings.}
% \label{fig:rq4_source_sensitivity}
% \end{figure}

\begin{figure}[!t]
\centering
\begin{subfigure}[t]{0.48\columnwidth}
\centering
\includegraphics[width=\linewidth]{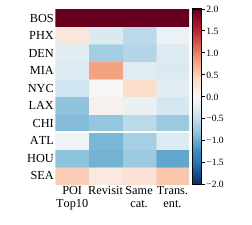}
\caption{City behavior}
\end{subfigure}
\hfill
\begin{subfigure}[t]{0.48\columnwidth}
\centering
\includegraphics[width=\linewidth]{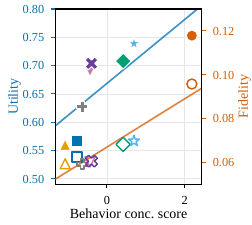}
\caption{Behavior concentration}
\end{subfigure}
\caption{City behavior priors and panel behavior concentration.}
\label{fig:rq4_source_sensitivity_bottom}
\end{figure}

% The lower panels provide a behavioral account of this sensitivity.
Figure~\ref{fig:rq4_source_sensitivity_bottom} (a) summarizes POI Top10 concentration, revisit rate, same-category persistence, and transition concentration, represented by negative normalized transition entropy.
Seattle is more concentrated than Houston across these priors.
This contrast is consistent with Figure~\ref{fig:rq4_source_sensitivity_top}: diffuse source mixtures reduce repeated structure more sharply for Seattle, lowering utility while mitigating over-concentration and improving fidelity.
Figure~\ref{fig:rq4_source_sensitivity_bottom} (b) reveals the same trade-off in the generated traces, where greater behavioral concentration is associated with higher utility but also larger fidelity JSD. Hence, adding more sources is not inherently beneficial. Across the tested mixtures, allocating 25--50\% of the activity budget to a relatively concentrated anchor and the remainder to a diverse background provides a practical compromise when target-region HAT statistics are unavailable.

\subsection{RQ5: Generation Efficiency}
% We measure efficiency by end-to-end generation throughput, defined as the number of generated traces per wall-clock second.
% We report two controlled regimes on one B200 node: a same-batch serial setting with one city worker and $512$ traces per city, and a maximum-GPU setting selected from a worker-batch sweep up to eight concurrent city workers.

\begin{table}[h]
\centering
\small
\caption{Generation efficiency with one and eight parallel workers. Higher throughput is better; lower memory is better.}
\label{tab:efficiency}
\resizebox{\columnwidth}{!}{%
\begin{tabular}{l c c c}
\hline
Method & 1-worker traces/s & 8-worker traces/s & Peak mem. (GB) \\
\hline
Markov & 159.36 & 920.38 & -- \\
TimeGeo & 22.86 & 127.35 & -- \\
LSTM & 13.48 & 22.78 & 48.52 \\
MobFormer & 12.08 & 16.19 & 39.46 \\
COLA & 13.31 & 18.68 & 40.73 \\
MobTA & 12.12 & 16.93 & 43.69 \\
ActivityEditor & 13.04 & 17.88 & 42.48 \\
\textbf{\m} & 18.39 & 67.26 & 6.74 \\
\hline
\end{tabular}
}
\end{table}

Table~\ref{tab:efficiency} compares end-to-end generation efficiency using one and eight parallel workers. \m generates $18.39$ traces per second with one worker and $67.26$ with eight, outperforming all neural baselines in both settings. With eight workers, it is approximately $3\times$ faster than the fastest neural baseline. \m uses only $6.74$ GB of peak GPU memory, compared with $39.46$--$48.52$ GB for the neural baselines. Most latency arises from consistency-aware autoregressive generation, while behavioral cloning, dynamic grounding, and PoE realization introduce limited overhead. Markov and TimeGeo remain faster as lightweight CPU samplers but achieve substantially lower utility and aggregate fidelity, as shown in Tables~\ref{tab:utility} and~\ref{tab:fidelity}.
Overall, \m achieves the best efficiency--quality trade-off among the neural methods.

\section{Related Work}
\subsection{Synthetic HAT Generation}

Although various methods have been proposed for HAT generation, most focus on \emph{single-region generation}~\citep{gong2025stcdm,geogen2026,xu2026synhat}, where models are trained and evaluated using HATs from the same region. These methods typically learn region-specific POI representations and transition patterns, preventing direct transfer to a new region with a distinct POI space and behavioral distribution, particularly when target-region HATs are unavailable for retraining or adaptation.
Recent efforts have explored generation or transfer across unseen regions, but under different settings. MobTA generates coarse-grained grid trajectories rather than semantically rich POI-level HATs~\citep{mobta2026}. COLA supports cross-region transfer but still relies on target-specific parameters and frequency priors estimated from target trajectories~\citep{cola2024}. In contrast, our study generates timestamped activities without target-region HATs or POI-level transition statistics.

\subsection{Behavior-Aware Human Activity Modeling}
Human activity is strongly shaped by recurring behaviors rather than random movement. Individuals repeatedly return to a small set of familiar or anchor locations, balance exploration with preferential return, and exhibit both frequency- and recency-driven revisitation patterns~\citep{alessandretti2018evidence, pappalardo2015returners, song2010modelling, gonzalez2008understanding}. Preserving these behavioral regularities is essential for generating realistic HATs.
Existing methods~\cite{cola2024, gong2025stcdm, xu2026synhat} typically learn them from city-specific POI sequences. Such representations entangle behavioral patterns with POI identities, limiting cross-region transfer. In contrast, \m learns approximately region-invariant actions for revisiting recent, frequent, or anchor POIs and dynamically grounds them onto compatible target-region POIs. Consistency and transition evidence further constrain this grounding, enabling behavior-aware generation without target-region HATs.

\section{Conclusion}

In this paper, we propose \m, a behavior-conditioned framework for zero-shot HAT generation without using authentic individual HATs in target regions. \m introduces three key technical components: (i) a multidimensional consistency-aware intent extractor that captures temporal, semantic, and relative spatial intents from the generated prefix; (ii) a cross-region behavioral cloning module that learns approximately region-invariant actions for revisiting recent, frequent, or anchor POIs with an abstention option; and (ii) a behavior-conditioned activity realization module that dynamically grounds these actions onto compatible target-region POIs. Together, these components enable behavioral transfer without explicit cross-region POI alignment. Extensive experiments on a ten-city benchmark show that \m achieves $4.5$--$6.4\times$ the downstream utility of existing methods and reduces generation-fidelity JSD by $15.6\%$--$40.8\%$. Ablation studies show that behavioral cloning drives utility, whereas consistency and transition evidence help constrain spatial drift. Source-composition experiments further reveal a target-dependent utility--fidelity trade-off. Moreover, \m achieves approximately $3\times$ the throughput of the fastest neural baseline while using substantially less peak GPU memory.

% We studied zero-shot HAT generation over unique region contexts without target-region HATs.
% \m separates transferable behavior from region-specific realization through consistency-aware prefix encoding, cross-region behavioral cloning, and behavior-conditioned activity realization.
% The dynamic action--POI graph maps region-invariant behavioral decisions to variable target POI sets, while PoE realization reconciles behavioral, consistency, and transition evidence.
% Across three unseen target regions, \m achieved $4.5$--$6.4\times$ the utility of the strongest adapted baseline and reduced average fidelity JSD by $15.6\%$--$40.8\%$.
% Ablations showed that behavioral cloning drives utility, whereas consistency and transition evidence constrain spatial drift; source-composition experiments further revealed a target-dependent utility--fidelity tradeoff.
% The model also reached $3\times$ the throughput of the fastest neural baseline with substantially lower peak GPU memory.
% Together, these results establish behavior-conditioned realization as an effective route to zero-shot HAT generation.
% Our evaluation remains limited to U.S. regions from one data platform and broader geographic validation and automatic source selection from public target context remain important directions.

\FloatBarrier
\bibliography{main}

@inproceedings{geogen2026,
  title = {GeoGen: A Two-stage Coarse-to-Fine Framework for Fine-grained Synthetic Location-based Social Network Trajectory Generation},
  author = {Xu, Rongchao and Cai, Kunlin and Jiang, Lin and Hong, Zhiqing and Tian, Yuan and Wang, Guang},
  booktitle = {Proceedings of the AAAI Conference on Artificial Intelligence},
  volume = {40},
  pages = {1373--1381},
  year = {2026},
  doi = {10.1609/aaai.v40i2.37111}
}

@article{xu2026synhat,
  title={SynHAT: A Two-stage Coarse-to-Fine Diffusion Framework for Synthesizing Human Activity Traces},
  author={Xu, Rongchao and Jiang, Lin and Yu, Dahai and Li, Ximiao and Wang, Guang},
  journal={Proceedings of the ACM on Interactive, Mobile, Wearable and Ubiquitous Technologies},
  volume={10},
  number={2},
  pages={1--35},
  year={2026},
  publisher={ACM New York, NY, USA}
}

@misc{deweydata2026,
  title = {Dewey: Academic Research Data},
  author = {{Dewey}},
  year = {2026},
  howpublished = {\url{https://www.deweydata.io/}},
  note = {Accessed: 2026-06-18}
}

@article{mobta2026,
  title = {Bus-Conditioned Zero-Shot Trajectory Generation via Task Arithmetic},
  author = {Liu, Shuai and Cao, Ning and Chen, Yile and Jiang, Yue and Cong, Gao},
  journal = {arXiv preprint arXiv:2602.13071},
  year = {2026}
}

@article{activityeditor2026,
  title = {ActivityEditor: Learning to Synthesize Physically Valid Human Mobility},
  author = {Yang, Chenjie and Jiang, Yutian and Liang, Anqi and Qi, Wei and Wu, Chenyu and Zhang, Junbo},
  journal = {arXiv preprint arXiv:2604.05529},
  year = {2026}
}

@article{cola2024,
  title = {COLA: Cross-city Mobility Transformer for Human Trajectory Simulation},
  author = {Wang, Yu and Zheng, Tongya and Liang, Yuxuan and Liu, Shunyu and Song, Mingli},
  journal = {arXiv preprint arXiv:2403.01801},
  year = {2024}
}

@inproceedings{gambs2012next,
  title={Next place prediction using mobility markov chains},
  author={Gambs, S{\'e}bastien and Killijian, Marc-Olivier and del Prado Cortez, Miguel N{\'u}{\~n}ez},
  booktitle={Proceedings of the first workshop on measurement, privacy, and mobility},
  pages={1--6},
  year={2012}
}

@article{vaswani2017attention,
  title={Attention is all you need},
  author={Vaswani, Ashish and Shazeer, Noam and Parmar, Niki and Uszkoreit, Jakob and Jones, Llion and Gomez, Aidan N and Kaiser, {\L}ukasz and Polosukhin, Illia},
  journal={Advances in neural information processing systems},
  volume={30},
  year={2017}
}

@article{hochreiter1997long,
  title={Long short-term memory},
  author={Hochreiter, Sepp and Schmidhuber, J{\"u}rgen},
  journal={Neural computation},
  volume={9},
  number={8},
  pages={1735--1780},
  year={1997},
  publisher={MIT press}
}

@article{timegeo2016,
  title = {The {TimeGeo} modeling framework for urban mobility without travel surveys},
  author = {Jiang, Shan and Yang, Yingxiang and Gupta, Siddharth and Veneziano, Daniele and Athavale, Shounak and Gonz\'{a}lez, Marta C.},
  journal = {Proceedings of the National Academy of Sciences},
  volume = {113},
  number = {37},
  pages = {E5370--E5378},
  year = {2016}
}

@inproceedings{deepmove2018,
  title = {DeepMove: Predicting Human Mobility with Attentional Recurrent Networks},
  author = {Feng, Jie and Li, Yong and Zhang, Chao and Sun, Funing and Meng, Fanchao and Guo, Ang and Jin, Depeng},
  booktitle = {Proceedings of the 2018 World Wide Web Conference},
  pages = {1459--1468},
  year = {2018}
}

@inproceedings{stan2021,
  title = {STAN: Spatio-Temporal Attention Network for Next Location Recommendation},
  author = {Luo, Yingtao and Liu, Qiang and Liu, Zhaocheng},
  booktitle = {Proceedings of The Web Conference 2021},
  year = {2021},
  doi = {10.1145/3442381.3449998}
}

@inproceedings{getnext2022,
  title = {GETNext: Trajectory Flow Map Enhanced Transformer for Next POI Recommendation},
  author = {Yang, Song and Liu, Jiamou and Zhao, Kaiqi},
  booktitle = {Proceedings of the 45th International ACM SIGIR Conference on Research and Development in Information Retrieval},
  year = {2022},
  doi = {10.1145/3477495.3531983}
}

@article{starhit2022,
  title = {Hierarchical Transformer with Spatio-Temporal Context Aggregation for Next Point-of-Interest Recommendation},
  author = {Xie, Jiayi and Chen, Zhenzhong},
  journal = {arXiv preprint arXiv:2209.01559},
  year = {2022}
}

@article{gong2025stcdm,
  title={STCDM: spatio-temporal contrastive diffusion model for check-in sequence generation},
  author={Gong, Letian and Guo, Shengnan and Lin, Yan and Liu, Yichen and Zheng, Erwen and Shuang, Yiwei and Lin, Youfang and Hu, Jilin and Wan, Huaiyu},
  journal={IEEE Transactions on Knowledge and Data Engineering},
  volume={37},
  number={4},
  pages={2141--2154},
  year={2025},
  publisher={IEEE}
}

@inproceedings{gu2024mamba,
  title={Mamba: Linear-time sequence modeling with selective state spaces},
  author={Gu, Albert and Dao, Tri},
  booktitle={First conference on language modeling},
  year={2024}
}

@article{gonzalez2008understanding,
  title={Understanding individual human mobility patterns},
  author={Gonzalez, Marta C and Hidalgo, Cesar A and Barabasi, Albert-Laszlo},
  journal={nature},
  volume={453},
  number={7196},
  pages={779--782},
  year={2008},
  publisher={Nature Publishing Group UK London}
}

@article{song2010modelling,
  title={Modelling the scaling properties of human mobility},
  author={Song, Chaoming and Koren, Tal and Wang, Pu and Barab{\'a}si, Albert-L{\'a}szl{\'o}},
  journal={Nature physics},
  volume={6},
  number={10},
  pages={818--823},
  year={2010},
  publisher={Nature Publishing Group UK London}
}

@article{pappalardo2015returners,
  title={Returners and explorers dichotomy in human mobility},
  author={Pappalardo, Luca and Simini, Filippo and Rinzivillo, Salvatore and Pedreschi, Dino and Giannotti, Fosca and Barab{\'a}si, Albert-L{\'a}szl{\'o}},
  journal={Nature communications},
  volume={6},
  number={1},
  pages={8166},
  year={2015},
  publisher={Nature Publishing Group UK London}
}

@article{alessandretti2018evidence,
  title={Evidence for a conserved quantity in human mobility},
  author={Alessandretti, Laura and Sapiezynski, Piotr and Sekara, Vedran and Lehmann, Sune and Baronchelli, Andrea},
  journal={Nature human behaviour},
  volume={2},
  number={7},
  pages={485--491},
  year={2018},
  publisher={Nature Publishing Group UK London}
}

\clearpage
\appendix
\section{Appendix}

\subsection{Dataset Details}
\label{app:dataset_details}
Table~\ref{tab:appendix_dataset_stats} summarizes the datasets used in our work. 
% Boston and Chicago serve as source cities, while the remaining eight cities are held-out targets. 
The benchmark is constructed from large-scale HATs across ten U.S. metropolitan areas provided by Dewey. Each city contains approximately one to one and a half million check-ins from hundreds of thousands of devices, providing substantial behavioral diversity while maintaining strict separation between source and target cities.

The cities also vary considerably in their urban opportunity spaces. The number of POIs ranges from $8.4$K in Boston to $51.5$K in Houston, while the number of semantic POI categories ranges from $167$ to $226$. This variation distinguishes our setting from standard within-city HAT generation, as the generator must transfer activity decisions learned from source cities to targets with different POI vocabularies, spatial extents, and category compositions. Temporal coverage also varies from $4$ to $25$ observed days, further testing whether synthetic traces preserve useful activity patterns under heterogeneous data coverage.

The coarse spatial-flow context is computed over geohash-5 cells. Each cell is approximately $4.9$ km high and $3.3$--$4.4$ km wide at the latitudes of the studied U.S. cities. The Cells column reports the number of occupied geohash-5 cells after preprocessing. These cells are used only to construct the coarse spatial-flow context and do not define the POI vocabulary. The length and width columns are derived from the bounding-box spans of available OSM POIs in the public-context profile.

\begin{table*}[t]
\centering
\small
\caption{HAT data statistics from 10 cities used in this paper.}
\label{tab:appendix_dataset_stats}
\begin{tabular}{l c c c c c c c c}
\hline
City & Visits & Devices & Days & POIs & Cells & Categories & Length km & Width km \\
\hline
Atlanta & 1.0M & 439.6K & 9 & 17.7K & 213 & 203 & 35.0 & 62.6 \\
Boston & 1.5M & 485.7K & 25 & 8.4K & 68 & 167 & 9.5 & 21.5 \\
Chicago & 1.5M & 556.8K & 8 & 38.2K & 98 & 211 & 42.1 & 34.5 \\
Denver & 1.0M & 409.8K & 8 & 17.3K & 127 & 210 & 45.6 & 38.8 \\
Houston & 1.0M & 410.3K & 4 & 51.5K & 249 & 226 & 66.8 & 57.8 \\
Los Angeles & 1.0M & 456.9K & 6 & 32.6K & 133 & 200 & 69.6 & 47.2 \\
Miami & 1.0M & 373.9K & 7 & 22.8K & 149 & 215 & 36.5 & 38.5 \\
New York & 1.5M & 513.6K & 11 & 29.9K & 103 & 185 & 46.9 & 47.1 \\
Phoenix & 1.0M & 439.0K & 8 & 23.2K & 153 & 213 & 48.3 & 30.3 \\
Seattle & 1.5M & 457.6K & 17 & 14.8K & 65 & 193 & 26.9 & 15.7 \\
\hline
\end{tabular}
\end{table*}

\begin{table*}[t]
\centering
\small
\caption{Source-behavior diagnostics for cities involved in the source-sensitivity analysis and the three representative target cities. POI Top10 is the visit mass of the ten most frequent POIs; Trans/10k is the number of unique POI transitions per 10K events.}
\label{tab:appendix_source_behavior}
\begin{tabular}{l l c c c c c c}
\hline
Split & City & POI Top10 & POI Ent. & Trans/10k & Trans Ent. & Seen-Before & Same-Cat \\
\hline
\multirow{7}{*}{Source}
& Boston & 0.1756 & 0.7790 & 4126.2 & 0.8464 & 0.1094 & 0.4627 \\
& Phoenix & 0.0847 & 0.8478 & 6717.6 & 0.9404 & 0.0532 & 0.2339 \\
& Denver & 0.0679 & 0.8543 & 6777.3 & 0.9458 & 0.0451 & 0.2309 \\
& Miami & 0.0663 & 0.8746 & 6556.3 & 0.9463 & 0.0750 & 0.2544 \\
& New York & 0.0619 & 0.8708 & 7114.9 & 0.9436 & 0.0589 & 0.2965 \\
& Los Angeles & 0.0470 & 0.8809 & 7191.5 & 0.9483 & 0.0608 & 0.2617 \\
& Chicago & 0.0448 & 0.8744 & 7554.7 & 0.9610 & 0.0436 & 0.2341 \\
\hline
\multirow{3}{*}{Target}
& Atlanta & 0.0740 & 0.8475 & 6388.7 & 0.9457 & 0.0409 & 0.2249 \\
& Houston & 0.0462 & 0.9069 & 8044.8 & 0.9708 & 0.0402 & 0.2197 \\
& Seattle & 0.0944 & 0.8173 & 5687.4 & 0.9186 & 0.0631 & 0.2913 \\
\hline
\end{tabular}
\end{table*}

\subsection{Baseline Details}
\label{app:baseline_details}
We adapt all baselines to the same zero-shot target-city interface used by \m.
Each method receives source-city HATs and target-city context, but does not use target-city HATs for training.

\begin{itemize}
    \item \textbf{Markov.} Markov~\cite{gambs2012next} estimates source-side transition dynamics over the trace state space and samples the next activity from these transition statistics. The sampled activity is then grounded to feasible target-city POIs using the same target context.
    \item \textbf{TimeGeo.} TimeGeo~\cite{timegeo2016} uses temporal and spatial mobility-law priors to generate plausible movement and visit timing. We adapt it by mapping the generated activity decisions to target-city POIs under the target POI and coarse-flow context.
    \item \textbf{LSTM.} LSTM~\cite{hochreiter1997long} is a recurrent sequence generator trained on source-city HATs. At inference time, its source-trained sequential policy is coupled with target-city POI grounding to produce timestamped target traces.
    \item \textbf{MobFormer.} MobFormer~\cite{vaswani2017attention} is a Transformer-style sequence generator trained on source traces. We use the same target-grounding interface to translate its source-learned sequential decisions into target-city POI visits.
    \item \textbf{COLA.} COLA \citep{cola2024} is adapted as a contextual location-generation baseline. It uses target POI and context features during grounding, while following the same zero-shot restriction that target-city HATs are never used for training.
    \item \textbf{MobTA.} MobTA \citep{mobta2026} is adapted as a neural mobility-transfer baseline. We train and apply its activity policy under the same source-data and target-context budget, then output timestamped target-city POI traces.
    \item \textbf{ActivityEditor.} ActivityEditor \citep{activityeditor2026} is adapted as an editing-based mobility generator. In our setting, it edits source-derived behavioral templates and realizes the edited sequence in the target-city POI space.
\end{itemize}

\subsection{Metric Details}
\label{app:metric_details}
This section introduces the utility and fidelity metrics used in the evaluation.
All metrics are computed separately for each target city and then aggregated only when a table or figure explicitly reports an average.

\paragraph{Utility.}
Utility evaluates whether synthetic target-city HATs can replace real target-city HATs for downstream applications.
For each target city and each downstream application model $m$, we train it on generated traces $\widehat{\mathcal{H}}_{c_t}$ and again on the real target-city traces $\mathcal{H}_{c_t}^{\mathrm{train}}$.
Both of the trained models are tested on the same held-out real target-city traces.
Let $q_{m,c_t}^{\mathrm{syn}}$ be a raw next-POI recommendation performance score after training on generated traces, and let $q_{m,c_t}^{\mathrm{real}}$ be the corresponding score after training on real target-city traces.
The normalized utility ratio is
\begin{equation}
    U_{m,q}(c_t) = \frac{q_{m,c_t}^{\mathrm{syn}}}{q_{m,c_t}^{\mathrm{real}} + \epsilon},
\end{equation}
where $q$ can be Recall@$10$ or another raw ranking metric and $\epsilon$ is a small numerical constant.
A ratio close to $1$ means that synthetic HATs support the downstream application almost as well as real target-city training HATs under the same model and test split.
The main paper reports Recall@$10$ utility ratios, and the extended evaluation additionally reports Recall@$1$, Recall@$5$, Recall@$20$, MRR@$20$, and NDCG@$10$.

Given a test instance $j$, let $y_j$ be the true next POI and let $\pi_j$ be the ranked list of predicted POIs.
Recall@$k$ is the fraction of test instances where $y_j$ appears in the top $k$ positions of $\pi_j$:
\begin{equation}
    \mathrm{Recall}@k = \frac{1}{n}\sum_{j=1}^{n}\mathbf{1}[y_j \in \pi_j^{1:k}].
\end{equation}
MRR@$20$ measures the reciprocal rank of the true next POI within the top $20$ candidates:
\begin{equation}
    \mathrm{MRR}@20 = \frac{1}{n}\sum_{j=1}^{n}\mathbf{1}[r_j \leq 20]\frac{1}{r_j},
\end{equation}
where $r_j$ is the one-indexed rank of $y_j$ in $\pi_j$.
NDCG@$10$ uses the same single-relevant-item next-POI setting and discounts lower ranks logarithmically:
\begin{equation}
    \mathrm{NDCG}@10 = \frac{1}{n}\sum_{j=1}^{n}\mathbf{1}[r_j \leq 10]\frac{1}{\log_2(r_j+1)}.
\end{equation}
The average utility column in the main table is the arithmetic mean of the four Recall@$10$ ratios from DeepMove, STAN, GetNext, and STARHIT.

\paragraph{Fidelity.}
Fidelity evaluates whether generated HATs match the distributional properties of held-out real target-city HATs.
For each target city, every trace is normalized into five aligned event channels: geohash-5 cell ID, POI ID, category ID, absolute-time ID, and inter-event-gap ID.
Special padding or invalid tokens are removed before constructing event and transition distributions.
For any two empirical distributions $P$ and $Q$, we use Jensen-Shannon divergence with base-$2$ logarithms:
\begin{equation}
    \mathrm{JSD}(P,Q)=\frac{1}{2}\mathrm{KL}(P\|M)+\frac{1}{2}\mathrm{KL}(Q\|M).
\end{equation}
Here $M=(P+Q)/2$.
Lower values indicate better fidelity, and the base-$2$ JSD is bounded by $[0,1]$.
We construct nine submetric distributions from these channels.
The marginal submetrics are Macro JSD over geohash-5 cells, POI JSD over POI IDs, Category JSD over semantic categories, Absolute-time JSD over time-of-day bins, Gap JSD over inter-event-gap bins, and Length JSD over trace lengths.
The transition submetrics are Macro-transition JSD, POI-transition JSD, and Category-transition JSD, each computed over adjacent event pairs within a trace.
The reported spatial fidelity in the main paper is POI JSD, because POI realization is the most fine-grained spatial output of the generator.
The reported temporal fidelity is the mean of absolute-time JSD, gap JSD, and length JSD.
The reported semantic fidelity is the mean of category JSD and category-transition JSD.
The reported average fidelity is the arithmetic mean of the reported spatial, temporal, and semantic fidelity groups.
In the four-axis benchmark reports, we also keep the raw diagnostic aggregates $\textit{fidelity mean JSD}$ and $\textit{transition mean JSD}$.
The former is the mean of Macro JSD, POI JSD, Category JSD, Absolute-time JSD, Gap JSD, and Length JSD.
The latter is the mean of Macro-transition JSD, POI-transition JSD, and Category-transition JSD.

\subsection{Extended Methodology Details}
\label{app:extended_methodology_details}

\subsubsection{Target Candidate Construction}
\label{app:target_candidate_construction}
At step $i$, candidate construction restricts the target POI catalog to a
compact realization set while retaining consistency-, context-, and
behavior-supported alternatives. We first score target POIs using
\begin{equation}
\begin{aligned}
R_i^{\mathrm{con}}(p)
&=s_i^{\mathrm{sp}}(p)
+\lambda_q\log P_i^{\mathrm{cat}}(q_p),\\
R_i^{\mathrm{ctx}}(p)
&=\lambda_x\phi_x(x_p,\mathcal{X}_{c_t})
+\lambda_f\log\!\left(\epsilon+
F_{c_t}(c_{\hat p_{i-1}},c_p)\right).
\end{aligned}
\end{equation}
The first score retrieves POIs consistent with the semantic and relative
spatial intent; the second favors public target popularity, category supply,
accessibility, and coarse flow. Let $\mathcal{P}_i^{\mathrm{int}}$ contain
POIs in the highest-scoring categories and regions decoded from $z_i$. We
construct
\begin{equation}
\begin{aligned}
\mathcal{K}_i^{\mathrm{con}}
&=\operatorname{TopK}_{k_{\mathrm{con}}}
  (\mathcal{P}_{c_t};R_i^{\mathrm{con}}),\\
\mathcal{K}_i^{\mathrm{ctx}}
&=\operatorname{TopK}_{k_{\mathrm{ctx}}}
  (\mathcal{P}_i^{\mathrm{int}};R_i^{\mathrm{ctx}}).
\end{aligned}
\end{equation}

The memory pool preserves POIs that can realize policy actions. For each
feasible $a_\kappa^b\in\mathcal{A}_i^+$, we query
$\mathcal{Q}_\kappa(\mathcal{M}_i,z_i)$ and rank its POIs by reverse visit
age for recent, visit count for frequent, and first-observed membership for
anchor. Thus,
\begin{equation}
\mathcal{K}_i^{\mathrm{mem}}
=\bigcup_{a_\kappa^b\in\mathcal{A}_i^+}
\operatorname{TopK}_{k_{\mathrm{mem}}}
\!\left(\mathcal{Q}_\kappa(\mathcal{M}_i,z_i);R_i^b\right).
\end{equation}
The final set is the deduplicated union
\begin{equation}
\mathcal{K}_i=\operatorname{Unique}\!\left(
\mathcal{K}_i^{\mathrm{con}}\cup
\mathcal{K}_i^{\mathrm{ctx}}\cup
\mathcal{K}_i^{\mathrm{mem}}\right).
\end{equation}
If the union exceeds the candidate budget, we retain the highest-ranked POI
from every feasible memory query and fill the remaining slots by
$S_i^{\mathrm{con}}$. An empty set falls back to the highest-scoring POI
under $R_i^{\mathrm{con}}$. All budgets are selected by source-side
validation and fixed across target regions. Construction uses only the
generated prefix and public target context, never target-region HATs or
target POI-transition counts.

\subsubsection{Training Details}
\label{app:training_details}
During source-only training, each source region $c\in\mathcal{C}_s$ serves as
the current realization region. Teacher forcing supplies the observed prefix
$h_{<i}$, and the observed next POI $p_i$ is added to $\mathcal{K}_i$ so that
all likelihood terms are defined. Let $y_i^{\mathrm{time}}$ and
$y_i^{\mathrm{gap}}$ be the discretized timestamp and interval labels.

Cloning labels are obtained by replaying the source trace. The actions that
recover the observed next POI are
\begin{equation}
\mathcal{R}_i(p_i)=\left\{a_\kappa^b\in\mathcal{A}_i^+:
g_{a_\kappa^b}(\mathcal{M}_i,z_i)=p_i\right\}.
\end{equation}
If $\mathcal{R}_i(p_i)$ is nonempty, we choose its first action under the
fixed priority $a_{\mathrm{cc}}^{\mathrm{rec}} \succ
a_{\mathrm{cc}}^{\mathrm{freq}} \succ a_{\mathrm{cc}}^{\mathrm{anc}} \succ
a_{\mathrm{cat}}^{\mathrm{rec}} \succ a_{\mathrm{cat}}^{\mathrm{freq}}$;
otherwise, $a_i^\star=a^{\mathrm{keep}}$.

For each source step, the component losses are
\begin{equation}
\begin{aligned}
\ell_i^{\mathrm{poi}}
&=-\log P_i^{\mathrm{real}}(p_i),\\
\ell_i^{\mathrm{time}}
&=-\log P_i^{\mathrm{time}}(y_i^{\mathrm{time}}),\\
\ell_i^{\mathrm{gap}}
&=-\log P_i^{\mathrm{gap}}(y_i^{\mathrm{gap}}),\\
\ell_i^{\mathrm{cat}}
&=-\log P_i^{\mathrm{cat}}(q_{p_i}),\\
\ell_i^{\mathrm{BC}}
&=-\log\pi_\theta(a_i^\star\mid u_i).
\end{aligned}
\end{equation}
The relative spatial score is trained by candidate-normalized
cross-entropy,
\begin{equation}
\ell_i^{\mathrm{sp}}=-\log
\frac{\exp s_i^{\mathrm{sp}}(p_i)}
{\sum_{p\in\mathcal{K}_i}\exp s_i^{\mathrm{sp}}(p)}.
\end{equation}
The complete objective is
\begin{equation}
\begin{split}
\mathcal{L}={}&\sum_{c\in\mathcal{C}_s}
\sum_{h\in\mathcal{H}_c}\sum_{i=1}^{|h|}
\Bigl(\ell_i^{\mathrm{poi}}
+\lambda_t\ell_i^{\mathrm{time}}
+\lambda_\Delta\ell_i^{\mathrm{gap}}\\
&+\lambda_q\ell_i^{\mathrm{cat}}
+\lambda_s\ell_i^{\mathrm{sp}}
+\lambda_{\mathrm{BC}}\ell_i^{\mathrm{BC}}\Bigr).
\end{split}
\end{equation}
Candidate retrieval, feasibility masks, source category-transition
statistics, and coarse-flow priors are non-learned. Gradients from the POI
likelihood pass through the PoE, grounding kernel, behavioral policy, and
consistency encoder, while the auxiliary losses directly supervise their
corresponding outputs. All neural parameters are jointly optimized; loss
weights, candidate budgets, and checkpoints are selected on held-out source
validation and then fixed for every target region. No target-region HAT is
used for optimization or model selection.

\subsubsection{Algorithms}
\label{app:model_algorithms}
Algorithms~\ref{alg:source_training} and~\ref{alg:inference} summarize the
source-only training and zero-shot generation procedures, respectively.

\begin{algorithm}[H]
\small
\caption{Source-only training of \m}
\label{alg:source_training}
\begin{algorithmic}[1]
\Require Source HATs and contexts
$\{(\mathcal{H}_c,\mathcal{X}_c):c\in\mathcal{C}_s\}$; action space
$\mathcal{A}$
\Ensure Trained neural parameters $\Theta$
\State Initialize $\Theta$
\For{each source minibatch $\mathcal{B}$}
    \ForAll{$(c,h,i)\in\mathcal{B}$}
        \State Encode $h_{<i}$ and predict consistency intent $z_i$
        \State Build $\mathcal{M}_i$, feasibility $f_i$, and cloning label
        $a_i^\star$
        \State Evaluate $\pi_\theta(\cdot\mid u_i)$
        \State Build $\mathcal{K}_i$ from $\mathcal{X}_c$ and include $p_i$
        \State Compute $P_i^{\mathrm{con}}$ from intent and source context
        \State Build $\mathcal{G}_i$ and compute $T_i$ and
        $P_i^{\mathrm{beh}}$
        \State Compute $S_i^{\mathrm{trans}}$ and PoE distribution
        $P_i^{\mathrm{real}}$
        \State Accumulate the weighted per-step loss
    \EndFor
    \State Update $\Theta$ by gradient descent
\EndFor
\State Select weights, budgets, and checkpoint on source validation
\State \Return $\Theta$
\end{algorithmic}
\end{algorithm}

Algorithm~\ref{alg:source_training} treats candidate sets and graph masks as
fixed within each step, while jointly updating the consistency encoder,
action policy, edge scorer, and realization parameters. Source validation is
the only signal used for model selection.

\begin{algorithm}[H]
\small
\caption{Zero-shot target-region HAT generation}
\label{alg:inference}
\begin{algorithmic}[1]
\Require Source-trained $\Theta$, source category transitions
$P_{\mathcal{C}_s}^{\mathrm{cat}}$, target context $\mathcal{X}_{c_t}$,
trace length $L$
\Ensure Synthetic target HAT $\hat h$
\State Initialize $\hat h_{<1}$ with a start token and initialize
$\mathcal{M}_1$
\For{$i=1$ to $L$}
    \State Encode $\hat h_{<i}$ and predict $z_i$
    \State Decode $\hat t_i$ from $P_i^{\mathrm{time}}$ and
    $P_i^{\mathrm{gap}}$
    \State Build $\mathcal{M}_i$, $f_i$, and
    $\pi_\theta(\cdot\mid u_i)$
    \State Construct $\mathcal{K}_i$ from $z_i$, $\mathcal{M}_i$, and
    $\mathcal{X}_{c_t}$
    \State Compute $P_i^{\mathrm{con}}$ over $\mathcal{K}_i$
    \State Build $\mathcal{G}_i$ and compute behavior grounding
    $P_i^{\mathrm{beh}}$
    \State Compute $S_i^{\mathrm{trans}}$ and PoE distribution
    $P_i^{\mathrm{real}}$
    \State Sample $\hat p_i\sim P_i^{\mathrm{real}}$
    \State Append $(\hat p_i,\hat t_i)$ and update prefix memory
\EndFor
\State \Return $\hat h$
\end{algorithmic}
\end{algorithm}

At inference, all quantities depend only on the generated prefix,
source-trained parameters and category dynamics, and publicly available target context.
The procedure never reads a target-region HAT or target POI-transition table.

\subsection{Extended Evaluation}
\label{app:extended_evaluation}
This section expands the four evaluation questions from the main text with the full metric breakdowns behind the reported summaries.
We report the complete source-sensitivity results, the full generation-fidelity submetrics, additional spatial and transition visualizations, and the extended downstream utility metrics for all four applications.
Each analysis paragraph precedes the table or figure it discusses.

\paragraph{Source sensitivity.}
Table~\ref{tab:appendix_source_sensitivity_full} reports the full target-city numbers behind Figure~\ref{fig:rq4_source_sensitivity_top}.
The single-source settings expose the utility--fidelity tension most clearly.
A concentrated Boston source gave the highest utility on all three targets, up to $0.8407$ on Houston, but its fidelity was the worst among all settings, reaching $0.1326$ JSD on Houston.
A more diffuse Chicago source reversed this pattern, with lower utility but better fidelity.
Adding more source cities continued this trend: the six-source pool gave the best or near-best fidelity, yet its utility dropped on Seattle from $0.6784$ under Boston to $0.4565$.
The two mixtures recovered most of the utility while keeping fidelity low.
The Boston-plus-six mixture reached $0.6507$ utility on Seattle at $0.0505$ JSD, and the balanced Boston--Chicago-plus-six mixture gave the best Houston utility of $0.8551$ at $0.0845$ JSD.
These numbers support the main-text rule that a concentrated anchor combined with a diverse source background retains transferable structure while improving fidelity robustness.

\begin{table*}[!ht]
\centering
\scriptsize
\caption{Complete source-sensitivity results on the three representative target cities. Utility is the average Recall@10 utility ratio over DeepMove and GetNext in this diagnostic; higher is better. Fidelity is mean JSD against held-out real target HATs; lower is better.}
\label{tab:appendix_source_sensitivity_full}
\resizebox{\textwidth}{!}{%
\begin{tabular}{l l rr rr rr}
\hline
\multirow{2}{*}{Setting} & \multirow{2}{*}{Sources} & \multicolumn{2}{c}{ATL} & \multicolumn{2}{c}{HOU} & \multicolumn{2}{c}{SEA} \\
\cline{3-8}
 & & Utility & Fidelity & Utility & Fidelity & Utility & Fidelity \\
\hline
BOS only & BOS & 0.7396 & 0.0871 & 0.8407 & 0.1326 & 0.6784 & 0.0677 \\
CHI only & CHI & 0.5669 & 0.0517 & 0.7636 & 0.0777 & 0.3695 & 0.0584 \\
BOS+CHI & BOS, CHI & 0.6946 & 0.0573 & 0.8512 & 0.0957 & 0.5779 & 0.0519 \\
CHI+DEN & CHI, DEN & 0.5944 & 0.0468 & 0.7315 & 0.0752 & 0.3539 & 0.0563 \\
4-source & BOS, CHI, DEN, PHX & 0.6666 & 0.0477 & 0.8063 & 0.0835 & 0.5986 & 0.0499 \\
6-source & BOS, CHI, DEN, LAX, NYC, PHX & 0.6346 & 0.0472 & 0.7904 & 0.0812 & 0.4565 & 0.0494 \\
BOS+6 mix & 0.50 BOS + 0.50 6-source & 0.7423 & 0.0602 & 0.8235 & 0.0989 & 0.6507 & 0.0505 \\
BOS+CHI+6 mix & 0.25 (BOS, CHI) + 0.75 6-source & 0.6927 & 0.0484 & 0.8551 & 0.0845 & 0.5639 & 0.0490 \\
\hline
\end{tabular}%
}
\end{table*}

\paragraph{Full generation fidelity.}
\label{app:full_fidelity}
Tables~\ref{tab:appendix_fidelity_atlanta}--\ref{tab:appendix_fidelity_seattle} report the complete submetric breakdown behind the three grouped fidelity scores in the main text.
The breakdown shows that \textsc{ZeroHAT} obtained the best average JSD on every reported city, at $0.1572$ on Atlanta, $0.1994$ on Houston, and $0.1496$ on Seattle, and that this advantage was broad rather than driven by a single channel.
It was best on the raw trajectory-distribution submetrics for travel distance, trace radius, and length, and best on cell, POI, and cell-transition JSD on Atlanta and Seattle.
The largest single margin appeared on POI JSD, where \textsc{ZeroHAT} reached $0.2259$ on Atlanta against $0.3971$ for the strongest baseline, which is consistent with the affordance-calibrated realization concentrating visits on plausible target POIs.
The remaining gaps also confirm two known limits.
POI-transition JSD stayed near $0.87$--$0.92$ for all methods, because an unobserved target POI vocabulary makes exact POI-pair transitions hard to match without target HATs.
On Houston, TimeGeo retained a small edge on category and category-transition JSD, matching the main-text observation that a strong mobility-law prior can fit coarse semantics on that city.

\begin{table*}[!ht]
\centering
\scriptsize
\caption{Full generation fidelity submetrics on Atlanta. Lower is better; \best{red} marks the best value in each column and \second{blue} marks the second-best value. Dist., Radius, and Len. are raw trajectory-distribution JSDs; the remaining columns are benchmark marginal and transition JSDs.}
\label{tab:appendix_fidelity_atlanta}
\resizebox{\textwidth}{!}{%
\begin{tabular}{l r r r r r r r r r r r}
\hline
Method & Dist. & Radius & Len. & Cell & POI & Cat. & Time & Cell-T & POI-T & Cat.-T & Avg. \\
\hline
Markov & 0.1808 & 0.1386 & 0.0314 & 0.0782 & 0.4086 & 0.0986 & 0.0000 & 0.3057 & 0.9932 & 0.2716 & 0.2507 \\
TimeGeo & \second{0.0482} & \second{0.0643} & 0.0302 & \second{0.0664} & \second{0.3971} & \second{0.0919} & 0.5336 & \second{0.1474} & 0.9876 & \second{0.2545} & 0.2621 \\
LSTM & 0.1880 & 0.1502 & 0.0348 & 0.0818 & 0.4156 & 0.1082 & \best{0.0000} & 0.3172 & 0.9940 & 0.2682 & 0.2558 \\
MobFormer & 0.1987 & 0.1542 & 0.0322 & 0.0810 & 0.4203 & 0.1150 & 0.0000 & 0.3249 & 0.9950 & 0.2822 & 0.2604 \\
COLA & 0.1970 & 0.1524 & 0.0340 & 0.0808 & 0.4127 & 0.1068 & \best{0.0000} & 0.3212 & 0.9940 & 0.2728 & 0.2572 \\
MobTA & 0.1524 & 0.1279 & 0.0399 & 0.0788 & 0.4147 & 0.1147 & \second{0.0000} & 0.2893 & 0.9943 & 0.2758 & \second{0.2488} \\
ActivityEditor & 0.1828 & 0.1423 & 0.0298 & 0.0791 & 0.4171 & 0.1124 & 0.5283 & 0.3096 & 0.9947 & 0.2803 & 0.3076 \\
\textsc{ZeroHAT} & \best{0.0191} & \best{0.0242} & \second{0.0089} & \best{0.0309} & \best{0.2259} & \best{0.0604} & \best{0.0000} & \best{0.0925} & \best{0.8896} & \best{0.2204} & \best{0.1572} \\
\hline
\end{tabular}%
}
\end{table*}

\begin{table*}[t]
\centering
\scriptsize
\caption{Full generation fidelity submetrics on Houston. Lower is better; \best{red} marks the best value in each column and \second{blue} marks the second-best value. Dist., Radius, and Len. are raw trajectory-distribution JSDs; the remaining columns are benchmark marginal and transition JSDs. Houston is the hardest target, with the largest POI vocabulary in Table~\ref{tab:appendix_dataset_stats}.}
\label{tab:appendix_fidelity_houston}
\resizebox{\textwidth}{!}{%
\begin{tabular}{l r r r r r r r r r r r}
\hline
Method & Dist. & Radius & Len. & Cell & POI & Cat. & Time & Cell-T & POI-T & Cat.-T & Avg. \\
\hline
Markov & 0.1233 & 0.0542 & 0.0792 & 0.0308 & 0.4867 & 0.0998 & \best{0.0000} & 0.3236 & 0.9980 & 0.2621 & 0.2458 \\
TimeGeo & \second{0.0165} & \best{0.0169} & 0.0790 & \best{0.0255} & \second{0.4687} & \best{0.0958} & 0.5352 & \best{0.1406} & 0.9953 & \best{0.2506} & 0.2624 \\
LSTM & 0.1280 & 0.0524 & 0.0802 & 0.0331 & 0.4876 & 0.1122 & \second{0.0000} & 0.3248 & 0.9988 & \second{0.2609} & 0.2478 \\
MobFormer & 0.1339 & 0.0521 & 0.0872 & 0.0313 & 0.4923 & 0.1141 & \best{0.0000} & 0.3343 & 0.9988 & 0.2676 & 0.2512 \\
COLA & 0.1287 & 0.0539 & 0.0841 & \second{0.0298} & 0.4847 & 0.1102 & \best{0.0000} & 0.3292 & 0.9984 & 0.2612 & 0.2480 \\
MobTA & 0.1037 & 0.0412 & 0.0879 & 0.0306 & 0.4810 & 0.1188 & 0.0000 & 0.2990 & 0.9983 & 0.2706 & \second{0.2431} \\
ActivityEditor & 0.1210 & 0.0496 & 0.0818 & 0.0313 & 0.4924 & 0.1188 & 0.5295 & 0.3203 & 0.9986 & 0.2735 & 0.3017 \\
\textsc{ZeroHAT} & \best{0.0050} & \second{0.0184} & \best{0.0484} & 0.0690 & \best{0.3763} & \second{0.0963} & \best{0.0000} & \second{0.1766} & \best{0.9222} & 0.2820 & \best{0.1994} \\
\hline
\end{tabular}%
}
\end{table*}

\begin{table*}[t]
\centering
\scriptsize
\caption{Full generation fidelity submetrics on Seattle. Lower is better; \best{red} marks the best value in each column and \second{blue} marks the second-best value. Dist., Radius, and Len. are raw trajectory-distribution JSDs; the remaining columns are benchmark marginal and transition JSDs.}
\label{tab:appendix_fidelity_seattle}
\resizebox{\textwidth}{!}{%
\begin{tabular}{l r r r r r r r r r r r}
\hline
Method & Dist. & Radius & Len. & Cell & POI & Cat. & Time & Cell-T & POI-T & Cat.-T & Avg. \\
\hline
Markov & 0.2098 & 0.2050 & 0.0398 & 0.0804 & 0.4469 & 0.0768 & \best{0.0000} & 0.3049 & 0.9928 & 0.2255 & 0.2582 \\
TimeGeo & 0.0634 & 0.0839 & 0.0404 & \second{0.0727} & \second{0.4459} & \second{0.0715} & 0.5379 & \second{0.1611} & 0.9885 & \second{0.2133} & 0.2679 \\
LSTM & 0.2172 & 0.2294 & 0.0428 & 0.0819 & 0.4656 & 0.0863 & \best{0.0000} & 0.3106 & 0.9951 & 0.2193 & 0.2648 \\
MobFormer & 0.2296 & 0.2110 & 0.0414 & 0.0843 & 0.4689 & 0.0925 & \best{0.0000} & 0.3240 & 0.9952 & 0.2344 & 0.2681 \\
COLA & 0.2300 & 0.2118 & 0.0469 & 0.0847 & 0.4703 & 0.0887 & \best{0.0000} & 0.3197 & 0.9948 & 0.2226 & 0.2669 \\
MobTA & 0.1858 & 0.1896 & 0.0491 & 0.0791 & 0.4711 & 0.0917 & 0.0000 & 0.2790 & 0.9943 & 0.2277 & \second{0.2567} \\
ActivityEditor & 0.2054 & 0.1913 & 0.0405 & 0.0810 & 0.4758 & 0.0917 & 0.5279 & 0.3044 & 0.9945 & 0.2339 & 0.3146 \\
\textsc{ZeroHAT} & \best{0.0208} & \best{0.0335} & \best{0.0168} & \best{0.0122} & \best{0.2359} & \best{0.0482} & \best{0.0000} & \best{0.0643} & \best{0.8740} & \best{0.1903} & \best{0.1496} \\
\hline
\end{tabular}%
}
\end{table*}

\paragraph{Additional fidelity visualizations.}
\label{app:additional_fidelity_visualization}
Figures~\ref{fig:appendix_spatial_density_atlanta}--\ref{fig:appendix_transition_matrix_seattle} present spatial-density and transition visualizations for the three representative target cities, extending the single-city results in the main text. Each figure includes Real and all compared generators except Markov, whose discrete transition model produces sparse patterns that are already substantially outperformed in the quantitative fidelity results.
Two consistent patterns emerge across cities. In the spatial-density maps, \m reproduces the Real concentration around major target-city POI clusters, whereas the neural baselines distribute visit mass more uniformly and fail to capture the dense cores. In the transition matrices, \m preserves multiple high-density within-category and cross-category blocks, while the baselines concentrate most probability mass in only a few coarse activity cells. These qualitative results help explain the spatial and transition JSD improvements reported in the fidelity tables.

\IfFileExists{figures/raw_hat_nature/fig_appendix_spatial_density_atlanta.pdf}{%
\begin{figure*}[!tp]
\centering
\includegraphics[width=\textwidth]{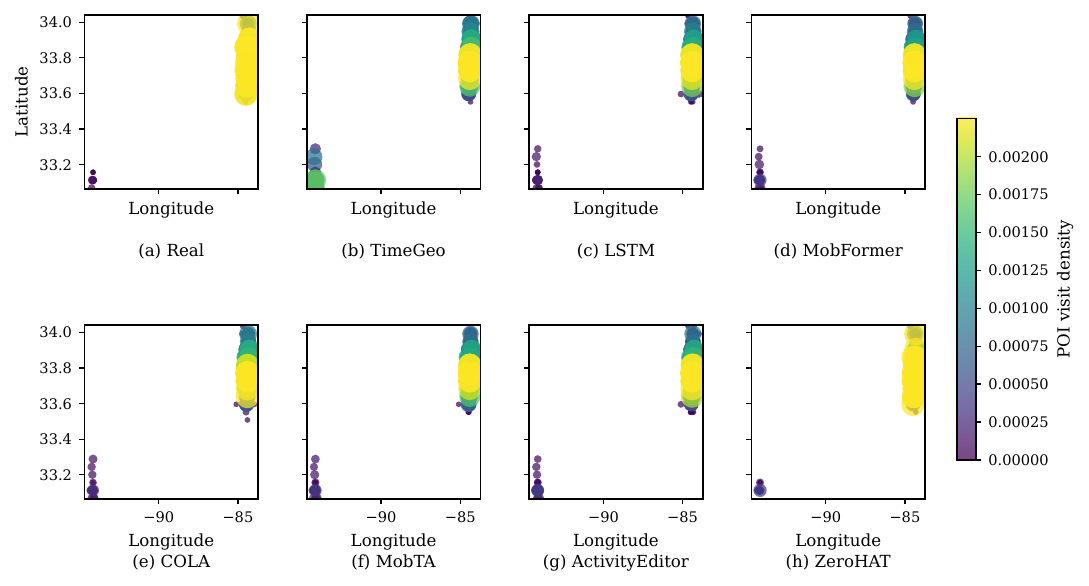}
\caption{POI visit-density scatter maps on Atlanta.}
\label{fig:appendix_spatial_density_atlanta}
\end{figure*}
}{}

\IfFileExists{figures/raw_hat_nature/fig_appendix_transition_matrix_atlanta.pdf}{%
\begin{figure*}[!tp]
\centering
\includegraphics[width=\textwidth]{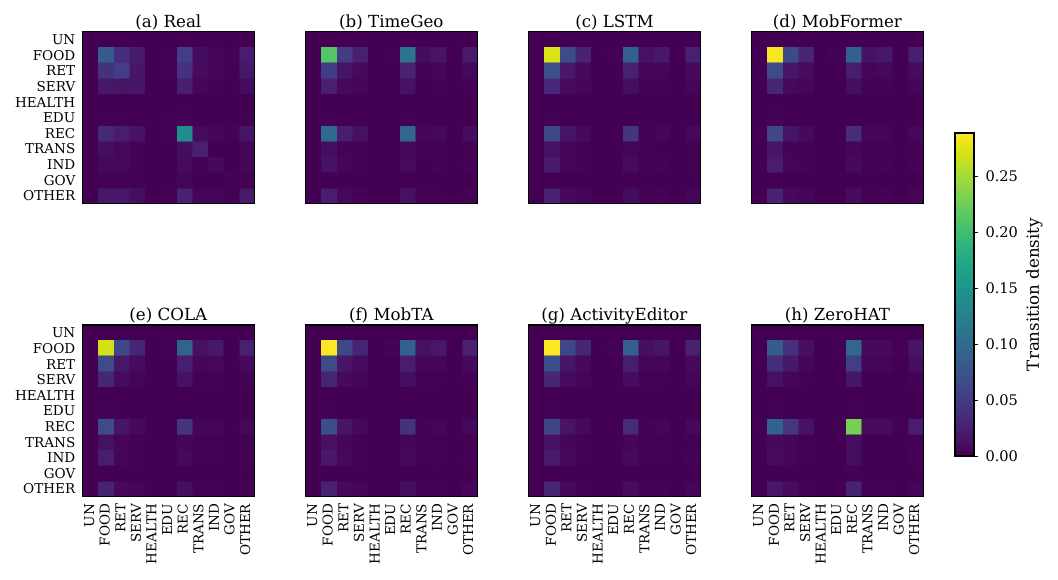}
\caption{Coarse-category transition density matrices on Atlanta.}
\label{fig:appendix_transition_matrix_atlanta}
\end{figure*}
}{}

\IfFileExists{figures/raw_hat_nature/fig_appendix_spatial_density_houston.pdf}{%
\begin{figure*}[!tp]
\centering
\includegraphics[width=\textwidth]{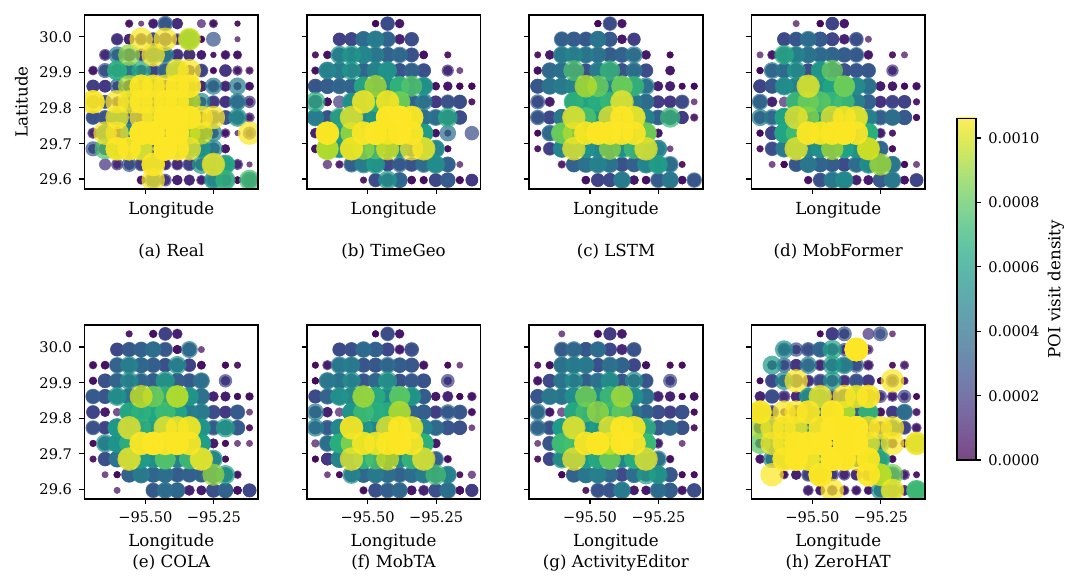}
\caption{POI visit-density scatter maps on Houston.}
\label{fig:appendix_spatial_density_houston}
\end{figure*}
}{}

\IfFileExists{figures/raw_hat_nature/fig_appendix_transition_matrix_houston.pdf}{%
\begin{figure*}[!tp]
\centering
\includegraphics[width=\textwidth]{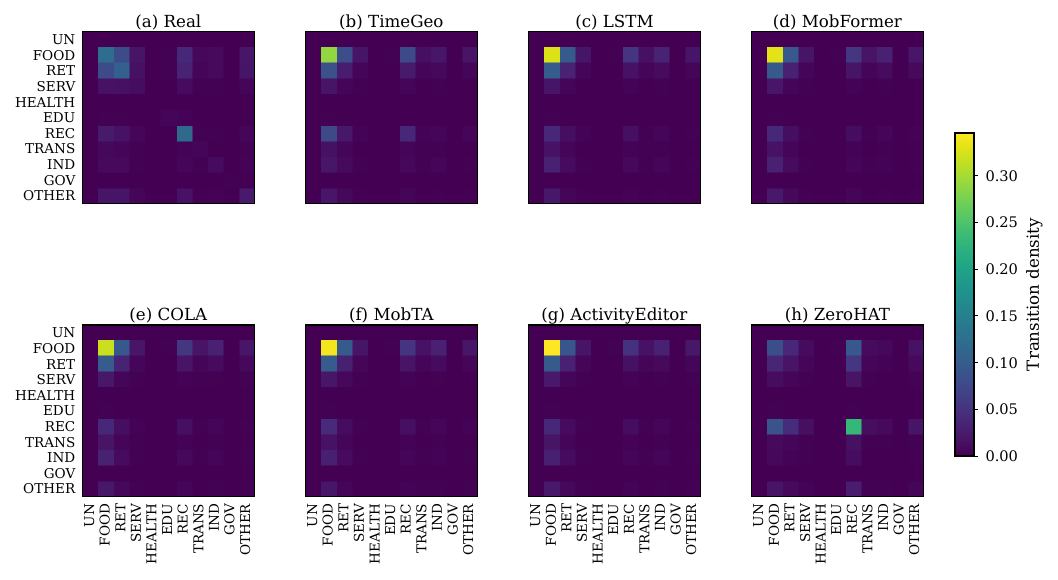}
\caption{Coarse-category transition density matrices on Houston.}
\label{fig:appendix_transition_matrix_houston}
\end{figure*}
}{}

\IfFileExists{figures/raw_hat_nature/fig_appendix_spatial_density_seattle.pdf}{%
\begin{figure*}[!tp]
\centering
\includegraphics[width=\textwidth]{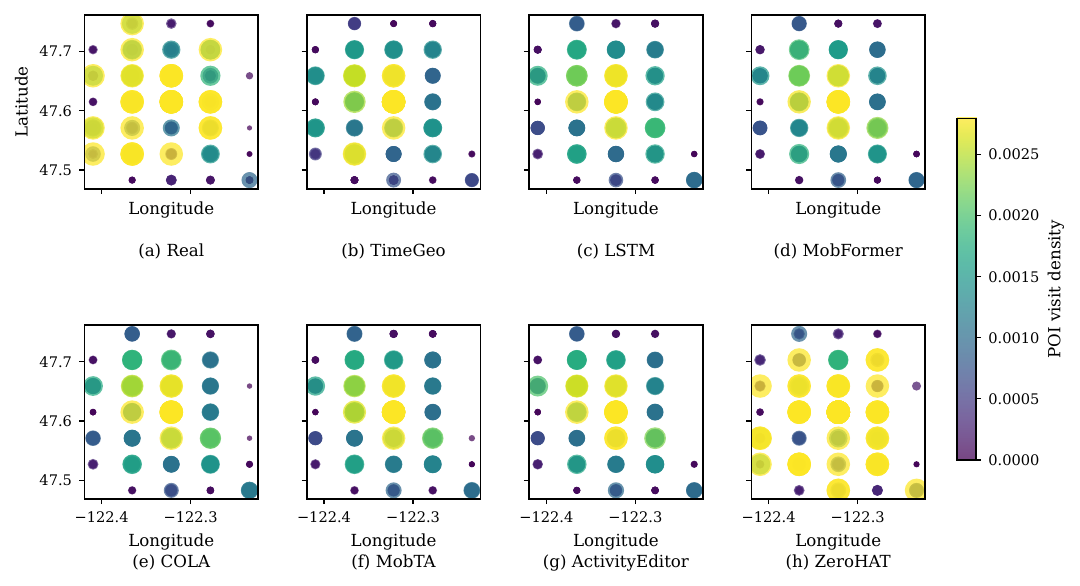}
\caption{POI visit-density scatter maps on Seattle. Each point denotes one POI, with size and color indicating normalized visit density.}
\label{fig:appendix_spatial_density_seattle}
\end{figure*}
}{%
\begin{figure*}[!tp]
\centering
\includegraphics[width=\textwidth]{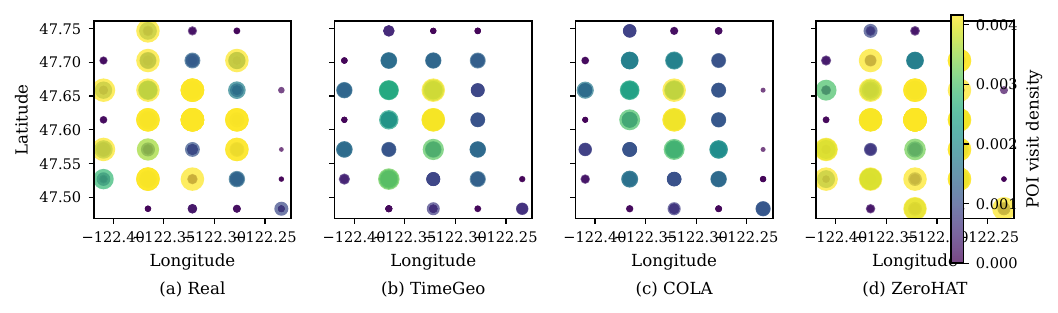}
\caption{POI visit-density scatter maps on Seattle. Each point denotes a POI, with size and color indicating normalized visit density.}
\label{fig:appendix_spatial_density_seattle}
\end{figure*}
}

\IfFileExists{figures/raw_hat_nature/fig_appendix_transition_matrix_seattle.pdf}{%
\begin{figure*}[!tp]
\centering
\includegraphics[width=\textwidth]{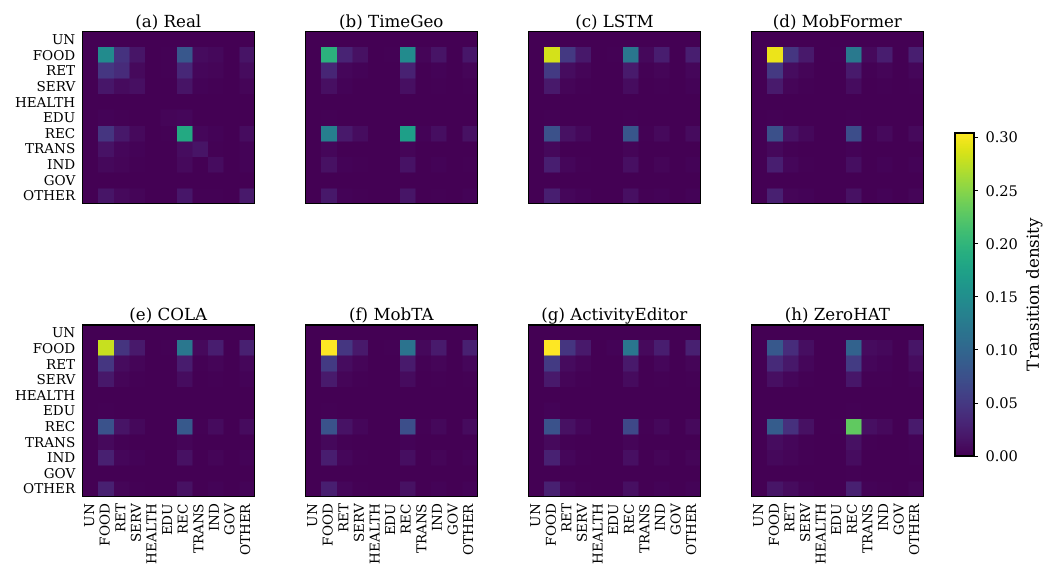}
\caption{Coarse-category transition density matrices on Seattle.}
\label{fig:appendix_transition_matrix_seattle}
\end{figure*}
}{}

\paragraph{Extended downstream utility.}
Tables~\ref{tab:extended_deepmove}--\ref{tab:extended_starhit} provide the full ranking-metric results for the four downstream applications, whereas the main text reports only the Recall@10 ratio. Each entry is a normalized utility ratio between a model trained on synthetic target-city HATs and the same model trained on real target-city HATs; higher is better.
The advantage of \m extends across all metrics, not only Recall@10. It achieves the best Recall@1, Recall@5, Recall@20, MRR@20, and NDCG@10 in every city--application setting, often by a large margin. The gains are especially pronounced at the strict Recall@1 cutoff. For GetNext on Houston, \m reaches $0.7985$ Recall@1, compared with $0.0311$ for the strongest baseline; for STARHIT on Houston, it achieves $0.7896$, compared with $0.0578$. These results indicate that the generated traces preserve the precise next-POI structure exploited by downstream rankers rather than merely matching coarser top-10 retrieval patterns. Baseline rankings remain broadly consistent across metrics, with TimeGeo and COLA typically performing best among the baseline models and the two source-only sequence models performing worst.

\begin{table*}[t]
\centering
\footnotesize
\caption{Extended utility metrics for DeepMove. All entries are normalized synthetic/original ratios. Higher is better; \best{red} marks the best value and \second{blue} the second-best value in each city-metric column.}
\label{tab:extended_deepmove}
\setlength{\tabcolsep}{4pt}
\begin{tabular}{l l r r r r r r}
\hline
City & Method & R@1 & R@5 & R@10 & R@20 & MRR@20 & NDCG@10 \\
\hline
Atlanta & Markov & 0.0323 & 0.0818 & 0.1095 & 0.1408 & 0.0634 & 0.0742 \\
 & TimeGeo & 0.0275 & 0.0820 & \second{0.1364} & \second{0.1684} & 0.0692 & \second{0.0858} \\
 & LSTM & 0.0105 & 0.0338 & 0.0677 & 0.0800 & 0.0313 & 0.0410 \\
 & MobFormer & 0.0072 & 0.0462 & 0.0593 & 0.1013 & 0.0345 & 0.0381 \\
 & COLA & \second{0.0379} & \second{0.0907} & 0.1226 & 0.1538 & \second{0.0731} & 0.0847 \\
 & MobTA & 0.0231 & 0.0571 & 0.0937 & 0.1506 & 0.0543 & 0.0608 \\
 & ActivityEditor & 0.0097 & 0.0437 & 0.0609 & 0.0815 & 0.0316 & 0.0385 \\
 & \m & \best{0.5404} & \best{0.5282} & \best{0.5225} & \best{0.5476} & \best{0.5342} & \best{0.5287} \\
\hline
Houston & Markov & 0.0894 & 0.1847 & 0.2079 & 0.2522 & 0.1498 & 0.1637 \\
 & TimeGeo & \second{0.0968} & 0.2049 & \second{0.2663} & \second{0.3224} & \second{0.1679} & \second{0.1932} \\
 & LSTM & 0.0346 & 0.0685 & 0.0988 & 0.1324 & 0.0595 & 0.0682 \\
 & MobFormer & 0.0202 & 0.0579 & 0.0759 & 0.1063 & 0.0470 & 0.0531 \\
 & COLA & 0.0789 & \second{0.2098} & 0.2385 & 0.3109 & 0.1583 & 0.1768 \\
 & MobTA & 0.0954 & 0.1757 & 0.2224 & 0.2531 & 0.1455 & 0.1656 \\
 & ActivityEditor & 0.0690 & 0.0967 & 0.1400 & 0.1771 & 0.0960 & 0.1057 \\
 & \m & \best{0.7888} & \best{0.7094} & \best{0.7464} & \best{0.7860} & \best{0.7564} & \best{0.7485} \\
\hline
Seattle & Markov & \second{0.0366} & 0.0627 & 0.0993 & 0.1509 & 0.0597 & 0.0668 \\
 & TimeGeo & 0.0191 & 0.0577 & 0.1187 & 0.1623 & 0.0503 & 0.0656 \\
 & LSTM & 0.0218 & 0.0404 & 0.0587 & 0.0931 & 0.0370 & 0.0406 \\
 & MobFormer & 0.0090 & 0.0218 & 0.0355 & 0.0572 & 0.0186 & 0.0216 \\
 & COLA & 0.0292 & \second{0.0980} & \second{0.1272} & \second{0.1765} & \second{0.0653} & \second{0.0789} \\
 & MobTA & 0.0329 & 0.0713 & 0.0990 & 0.1373 & 0.0595 & 0.0676 \\
 & ActivityEditor & 0.0141 & 0.0482 & 0.0653 & 0.0840 & 0.0342 & 0.0415 \\
 & \m & \best{0.4960} & \best{0.4769} & \best{0.5651} & \best{0.6324} & \best{0.5023} & \best{0.5133} \\
\hline
\end{tabular}
\end{table*}

\begin{table*}[t]
\centering
\footnotesize
\caption{Extended utility metrics for STAN. All entries are normalized synthetic/original ratios. Higher is better; \best{red} marks the best value and \second{blue} the second-best value in each city-metric column.}
\label{tab:extended_stan}
\setlength{\tabcolsep}{4pt}
\begin{tabular}{l l r r r r r r}
\hline
City & Method & R@1 & R@5 & R@10 & R@20 & MRR@20 & NDCG@10 \\
\hline
Atlanta & Markov & 0.0163 & 0.0502 & 0.0684 & 0.1180 & 0.0416 & 0.0449 \\
 & TimeGeo & 0.0201 & 0.0443 & 0.0791 & 0.1110 & 0.0424 & 0.0500 \\
 & LSTM & 0.0120 & 0.0324 & 0.0452 & 0.0663 & 0.0273 & 0.0307 \\
 & MobFormer & \second{0.0370} & 0.0327 & 0.0429 & 0.0575 & 0.0382 & 0.0382 \\
 & COLA & 0.0320 & 0.0532 & 0.0678 & 0.1048 & 0.0496 & 0.0517 \\
 & MobTA & 0.0279 & \second{0.0816} & \second{0.0964} & \second{0.1306} & \second{0.0592} & \second{0.0676} \\
 & ActivityEditor & 0.0078 & 0.0199 & 0.0356 & 0.0514 & 0.0184 & 0.0220 \\
 & \m & \best{0.4688} & \best{0.4843} & \best{0.5383} & \best{0.5559} & \best{0.4896} & \best{0.5015} \\
\hline
Houston & Markov & 0.0226 & 0.0576 & 0.0811 & 0.1152 & 0.0451 & 0.0528 \\
 & TimeGeo & \second{0.0370} & 0.0676 & 0.0965 & 0.1774 & 0.0645 & 0.0669 \\
 & LSTM & 0.0081 & 0.0562 & 0.0734 & 0.1101 & 0.0296 & 0.0394 \\
 & MobFormer & 0.0056 & 0.0206 & 0.0356 & 0.0523 & 0.0167 & 0.0208 \\
 & COLA & 0.0324 & 0.0699 & \second{0.1273} & 0.1734 & 0.0645 & 0.0784 \\
 & MobTA & 0.0253 & \second{0.0972} & 0.1264 & \second{0.1865} & \second{0.0667} & \second{0.0796} \\
 & ActivityEditor & 0.0096 & 0.0223 & 0.0417 & 0.0841 & 0.0231 & 0.0251 \\
 & \m & \best{0.7136} & \best{0.7083} & \best{0.7482} & \best{0.8121} & \best{0.7176} & \best{0.7206} \\
\hline
Seattle & Markov & 0.0207 & 0.0385 & 0.0671 & 0.1019 & 0.0363 & 0.0420 \\
 & TimeGeo & 0.0143 & 0.0385 & 0.0585 & 0.0907 & 0.0317 & 0.0368 \\
 & LSTM & 0.0051 & 0.0192 & 0.0334 & 0.0616 & 0.0163 & 0.0190 \\
 & MobFormer & 0.0052 & 0.0182 & 0.0278 & 0.0447 & 0.0133 & 0.0160 \\
 & COLA & \second{0.0516} & \second{0.0706} & \second{0.1000} & \second{0.1264} & \second{0.0648} & \second{0.0723} \\
 & MobTA & 0.0184 & 0.0448 & 0.0627 & 0.0981 & 0.0366 & 0.0413 \\
 & ActivityEditor & 0.0069 & 0.0311 & 0.0428 & 0.0653 & 0.0235 & 0.0274 \\
 & \m & \best{0.4943} & \best{0.4858} & \best{0.5664} & \best{0.6299} & \best{0.5013} & \best{0.5132} \\
\hline
\end{tabular}
\end{table*}

\begin{table*}[t]
\centering
\footnotesize
\caption{Extended utility metrics for GetNext. All entries are normalized synthetic/original ratios. Higher is better; \best{red} marks the best value and \second{blue} the second-best value in each city-metric column.}
\label{tab:extended_getnext}
\setlength{\tabcolsep}{4pt}
\begin{tabular}{l l r r r r r r}
\hline
City & Method & R@1 & R@5 & R@10 & R@20 & MRR@20 & NDCG@10 \\
\hline
Atlanta & Markov & 0.0191 & 0.0475 & 0.0684 & 0.0926 & 0.0399 & 0.0461 \\
 & TimeGeo & 0.0342 & 0.0461 & 0.0646 & 0.0778 & 0.0454 & 0.0498 \\
 & LSTM & 0.0150 & 0.0289 & 0.0506 & 0.0674 & 0.0279 & 0.0330 \\
 & MobFormer & 0.0054 & 0.0222 & 0.0444 & 0.0805 & 0.0218 & 0.0253 \\
 & COLA & \second{0.0474} & \second{0.0628} & \second{0.0844} & 0.1101 & \second{0.0624} & \second{0.0665} \\
 & MobTA & 0.0194 & 0.0448 & 0.0690 & \second{0.1134} & 0.0417 & 0.0458 \\
 & ActivityEditor & 0.0094 & 0.0271 & 0.0405 & 0.0632 & 0.0233 & 0.0264 \\
 & \m & \best{0.5928} & \best{0.5350} & \best{0.5334} & \best{0.5445} & \best{0.5631} & \best{0.5528} \\
\hline
Houston & Markov & \second{0.0311} & 0.0662 & 0.0925 & 0.1305 & 0.0557 & 0.0633 \\
 & TimeGeo & 0.0250 & \second{0.0766} & \second{0.1093} & \second{0.1491} & \second{0.0575} & \second{0.0694} \\
 & LSTM & 0.0067 & 0.0226 & 0.0388 & 0.0614 & 0.0198 & 0.0235 \\
 & MobFormer & 0.0048 & 0.0153 & 0.0297 & 0.0506 & 0.0151 & 0.0176 \\
 & COLA & 0.0216 & 0.0514 & 0.0754 & 0.1035 & 0.0429 & 0.0502 \\
 & MobTA & 0.0153 & 0.0407 & 0.0795 & 0.1203 & 0.0383 & 0.0466 \\
 & ActivityEditor & 0.0185 & 0.0375 & 0.0593 & 0.0996 & 0.0335 & 0.0378 \\
 & \m & \best{0.7985} & \best{0.7654} & \best{0.7976} & \best{0.8435} & \best{0.7922} & \best{0.7893} \\
\hline
Seattle & Markov & \second{0.0224} & 0.0533 & 0.0739 & 0.1062 & 0.0416 & 0.0483 \\
 & TimeGeo & 0.0204 & 0.0463 & 0.0825 & \second{0.1396} & 0.0420 & 0.0486 \\
 & LSTM & 0.0094 & 0.0264 & 0.0439 & 0.0607 & 0.0211 & 0.0260 \\
 & MobFormer & 0.0066 & 0.0199 & 0.0293 & 0.0447 & 0.0152 & 0.0180 \\
 & COLA & 0.0170 & 0.0392 & 0.0695 & 0.1123 & 0.0351 & 0.0413 \\
 & MobTA & 0.0158 & \second{0.0630} & \second{0.0947} & 0.1362 & \second{0.0459} & \second{0.0564} \\
 & ActivityEditor & 0.0219 & 0.0377 & 0.0544 & 0.0866 & 0.0334 & 0.0366 \\
 & \m & \best{0.5509} & \best{0.5262} & \best{0.5869} & \best{0.6672} & \best{0.5462} & \best{0.5504} \\
\hline
\end{tabular}
\end{table*}

\begin{table*}[t]
\centering
\footnotesize
\caption{Extended utility metrics for STARHIT. All entries are normalized synthetic/original ratios. Higher is better; \best{red} marks the best value and \second{blue} the second-best value in each city-metric column.}
\label{tab:extended_starhit}
\setlength{\tabcolsep}{4pt}
\begin{tabular}{l l r r r r r r}
\hline
City & Method & R@1 & R@5 & R@10 & R@20 & MRR@20 & NDCG@10 \\
\hline
Atlanta & Markov & 0.0202 & \second{0.0729} & 0.0988 & 0.1270 & 0.0513 & 0.0628 \\
 & TimeGeo & 0.0277 & 0.0641 & \second{0.1128} & \second{0.1477} & \second{0.0582} & \second{0.0707} \\
 & LSTM & 0.0070 & 0.0311 & 0.0409 & 0.0625 & 0.0219 & 0.0257 \\
 & MobFormer & 0.0051 & 0.0247 & 0.0495 & 0.0767 & 0.0233 & 0.0287 \\
 & COLA & 0.0165 & 0.0701 & 0.0984 & 0.1300 & 0.0497 & 0.0611 \\
 & MobTA & \second{0.0334} & 0.0666 & 0.0943 & 0.1245 & 0.0570 & 0.0651 \\
 & ActivityEditor & 0.0069 & 0.0336 & 0.0521 & 0.0688 & 0.0246 & 0.0312 \\
 & \m & \best{0.5675} & \best{0.5445} & \best{0.5308} & \best{0.5617} & \best{0.5536} & \best{0.5446} \\
\hline
Houston & Markov & 0.0431 & 0.1246 & 0.1501 & 0.1884 & 0.0871 & 0.1027 \\
 & TimeGeo & \second{0.0578} & \second{0.1597} & \second{0.1943} & \second{0.2401} & \second{0.1182} & \second{0.1371} \\
 & LSTM & 0.0146 & 0.0352 & 0.0559 & 0.0809 & 0.0306 & 0.0357 \\
 & MobFormer & 0.0456 & 0.0532 & 0.0636 & 0.0777 & 0.0531 & 0.0550 \\
 & COLA & 0.0517 & 0.1425 & 0.1675 & 0.2072 & 0.1040 & 0.1199 \\
 & MobTA & 0.0547 & 0.1021 & 0.1412 & 0.1744 & 0.0892 & 0.1017 \\
 & ActivityEditor & 0.0279 & 0.0754 & 0.0858 & 0.1083 & 0.0533 & 0.0612 \\
 & \m & \best{0.7896} & \best{0.7057} & \best{0.7128} & \best{0.7564} & \best{0.7446} & \best{0.7306} \\
\hline
Seattle & Markov & 0.0236 & 0.0575 & 0.0944 & 0.1308 & 0.0468 & 0.0571 \\
 & TimeGeo & 0.0319 & \second{0.0847} & \second{0.1113} & \second{0.1560} & \second{0.0630} & \second{0.0731} \\
 & LSTM & 0.0074 & 0.0279 & 0.0387 & 0.0594 & 0.0192 & 0.0231 \\
 & MobFormer & 0.0030 & 0.0158 & 0.0296 & 0.0531 & 0.0125 & 0.0154 \\
 & COLA & 0.0262 & 0.0534 & 0.0790 & 0.1276 & 0.0466 & 0.0520 \\
 & MobTA & 0.0204 & 0.0505 & 0.0762 & 0.1242 & 0.0423 & 0.0481 \\
 & ActivityEditor & \second{0.0329} & 0.0569 & 0.0982 & 0.1347 & 0.0526 & 0.0621 \\
 & \m & \best{0.5380} & \best{0.5610} & \best{0.6371} & \best{0.6868} & \best{0.5600} & \best{0.5767} \\
\hline
\end{tabular}
\end{table*}

\end{document}